\pdfoutput=1

\documentclass[11pt]{article}

\usepackage[preprint]{acl}

\usepackage{times}
\usepackage{latexsym}
\usepackage{ulem}
\usepackage{soulutf8}    %

\usepackage[T1]{fontenc}

\usepackage[utf8]{inputenc}

\usepackage{microtype}

\usepackage{inconsolata}

\usepackage{graphicx}
\usepackage{booktabs}      %
\usepackage{multirow}      %
\usepackage{array}         %
\usepackage{makecell} %

\title{\datasetname: A Benchmark for Evaluating Exploration-driven Decision-making in Virtual Escape Rooms}

\usepackage{amsmath}
\usepackage{amssymb}
\usepackage{float}
\usepackage{graphicx}
\usepackage{amsmath}
\usepackage{booktabs}
\usepackage{multicol}
\usepackage{multirow}
\usepackage{array}
\usepackage{xspace}
\usepackage{caption}     
\usepackage{subcaption}
\usepackage{tabularx}
\usepackage{graphicx}
\usepackage{array}
\usepackage{graphicx} %
\usepackage{tikz} %

\usepackage{colortbl} %
\usepackage{amsmath} %

\author{
\textbf{Seungwon Lim} \hspace{0.5cm}
\textbf{Sungwoong Kim} \hspace{0.5cm}
\textbf{Jihwan Yu} \hspace{0.5cm}
\textbf{Sungjae Lee} \hspace{0.5cm} \\ [0.1cm]
\textbf{Jiwan Chung} \hspace{0.5cm}
\textbf{Youngjae Yu} \textsuperscript{\textdagger} \\[0.2cm]
{Yonsei University}
\\
  \small{
    \href{mailto:email@domain}{sngwon@yonsei.ac.kr}
  }
}

\newcommand{\eg}{e.g.,\xspace}

\newcommand{\exphint}{\(exp_{\text{hint}}\)}
\newcommand{\expbase}{\(exp_{\text{base}}\)}
\newcommand{\datasetname}{VisEscape\xspace}
\newcommand{\modelname}{VisEscaper\xspace}

\newcommand{\claude}{\textit{Claude 3.5 Sonnet}}
\newcommand{\gpt}{\textit{GPT-4o}}
\newcommand{\gptmini}{\textit{GPT-4o-mini}}
\newcommand{\geminipro}{\textit{Gemini-1.5-Pro}}
\newcommand{\geminiflash}{\textit{Gemini-2.0-Flash}}
\newcommand{\qwenvl}{\textit{Qwen2.5VL-32B}}
\newcommand{\internfive}{\textit{InternVL2.5-38B}}
\newcommand{\interntwo}{\textit{InternVL2-40B}}
\newcommand{\llava}{\textit{LLaVA-v1.6-34B}}
\newcommand{\llavaov}{\textit{LLaVA-OneVision-7B}}

\newcommand{\gradientcell}[1]{%
  \begin{tikzpicture}[baseline,anchor=base west]
    \fill[gray!25] (0,0) rectangle (0.9,0.25);
    \fill[gray!75] (0,0) rectangle ({0.9*#1/50},0.25);
    \node[anchor=west] at (0.9,0.15) {#1};
  \end{tikzpicture}%
}

\newcommand{\github}{\raisebox{-1.5pt}{\includegraphics[height=1.05em]{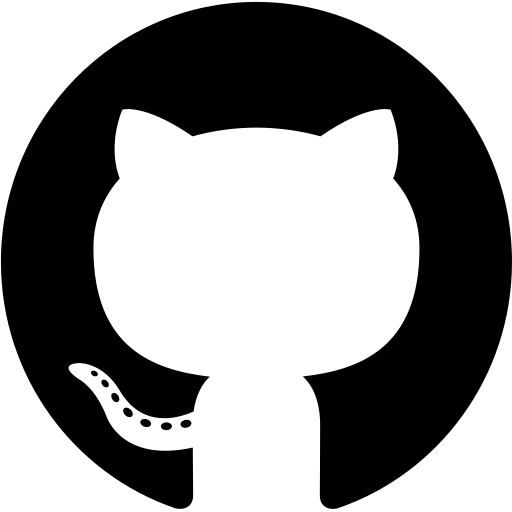}}\xspace}

\renewcommand{\and}{%
  \end{tabular}%
  \hskip 0.01em \begin{tabular}[t]{c}
}

\begin{document}
\maketitle
\begin{abstract}
Escape rooms present a unique cognitive challenge that demands exploration-driven planning: with the sole instruction to \textit{Escape the room}, players must actively search their environment, collecting information, and finding solutions through repeated trial and error. Motivated by this, we introduce \textbf{\datasetname}, a benchmark of 20 virtual escape rooms specifically designed to evaluate AI models under these challenging conditions, where success depends not only on solving isolated puzzles but also on iteratively constructing and refining spatial-temporal knowledge of a dynamically changing environment. On \datasetname, we observe that even state-of-the-art multi-modal models generally fail to escape the rooms, showing considerable variation in their progress and problem-solving approaches. We find that integrating memory management and reasoning contributes to efficient exploration and enables successive hypothesis formulation and testing, thereby leading to significant improvements in dynamic and exploration-driven environments\footnote{\github \textbf{Code\&Dataset}: \href{https://github.com/pull-ups/VisEscape}{pull-ups/VisEscape}}.

\end{abstract}

\section{Introduction}

Realistic goals often necessitate exploration:
when cooking with available ingredients, one must search through cabinets to identify ingredients and determine how to combine them into a meal.
Likewise, a trip to a foreign city cannot be planned without searching for major sights and train schedules.
In such cases, effective planning depends on acquiring information through interaction with the environment rather than having it provided a priori.

Existing works on embodied agents have addressed the importance of exploration, but it has primarily been studied in the context of decomposable and explicit tasks like ``Place a cleaned egg in a microwave''~\cite{Shridhar_2020_CVPR,puig2018virtualhome,li2021igibson,kim2024realfred}.
However, the capability of agents to autonomously search for and utilize relevant information in environments where tasks and solutions are not predefined has not been sufficiently investigated.

\begin{figure}[t]
\centering
\includegraphics[width=\linewidth]{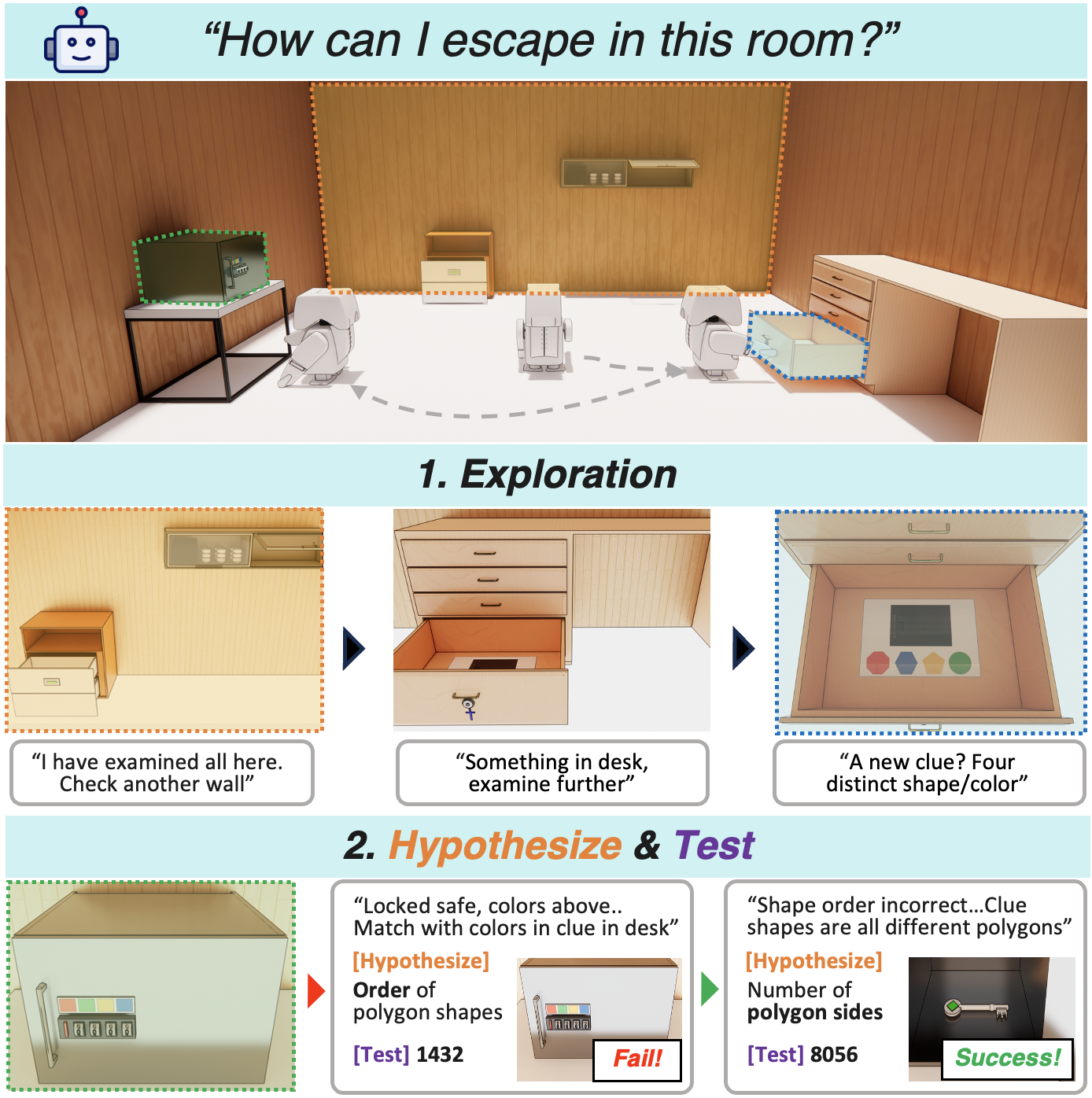}

\caption{Depiction of the exploration-driven problem-solving in \datasetname. Agents must (1) actively explore to uncover relevant information and (2) subsequently formulate and test hypotheses through interaction to solve puzzles to a successful escape.}
\label{fig:fig1}
\vspace{-3mm}

\end{figure}

\begin{figure*}[!htbp]
\centering
\includegraphics[width=\linewidth]{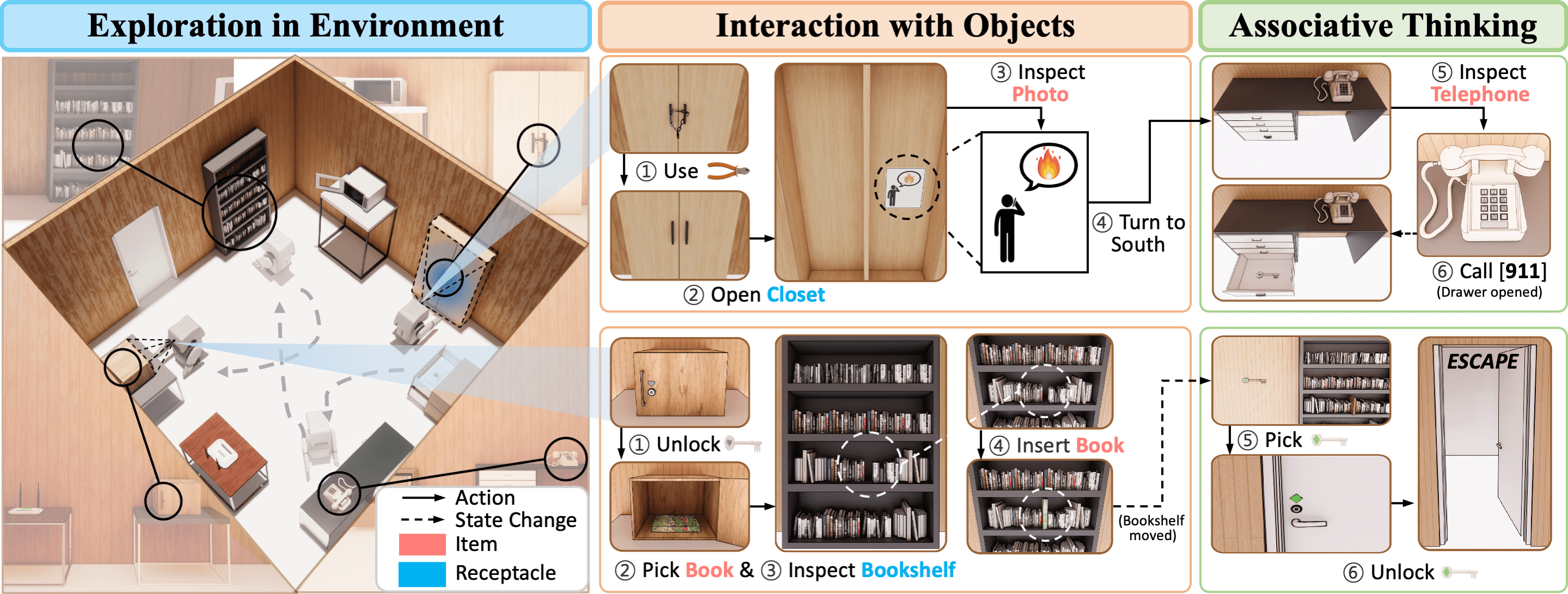}
\caption{An illustration of an excerpt of a trajectory from \datasetname. To escape the room successfully, agents must explore multiple directions and diverse views, and interact with various objects. Additionally, they need to infer associations between two or more scenes in different locations to solve creative puzzles.}
\label{fig:room_image}
\vspace{-5mm}

\end{figure*}

To bridge this gap, we argue that practical evaluation of decision-making agents should assess their ability to conduct self-directed exploration to identify underlying objectives and subsequently take reasoned actions. In particular, this entails formulating hypotheses for problem-solving based on information and clues gathered through exploration.

To evaluate agents under these conditions, we propose \datasetname, a benchmark of 20 virtual escape rooms for assessing exploration-centric problem-solving capacities of multimodal agents. Given the implicit goal of ``escape the room'', agents explore their surroundings, integrate relevant information, discover how to use objects, and solve complex puzzles to escape.  Moreover, they must retain previously observed information and interpret indirect clues, and reason beyond immediate observations while dynamically adapting to environmental changes. 

The evaluation of various MLLMs on \datasetname showed that even advanced models like Claude-3.5-Sonnet and GPT-4o exhibited low escape rates of less than 10\% without any hints. To test how memory and reasoning, which are core cognitive abilities for both agents and humans~\cite{}, navigate the cognitive challenges posed by \datasetname, we apply dedicated memory management and reasoning modules to models.

We find that this integration enables models that were previously wholly unsuccessful in escaping to achieve successful escapes; achieving x2.4 enhanced effectiveness and x3.8 improved efficiency. Through analysis, we observe that memory management reduces repetitive actions, enabling efficient exploration, and that memory and reasoning exhibit a synergistic effect.

\section{Key Challenges of \datasetname}
\label{sec:construction}
Figure~\ref{fig:room_image} illustrates an example of the problem-solving process required in the \datasetname. The game agent should execute a sequence of tasks comprising diverse interactions.

\textbf{ ~1. Self-directed Exploration.}
As in real escape room games, the game agent in \datasetname receives no instruction except for the instruction `escape the room'. Therefore, without a clear roadmap, the agent must actively explore and figure out how to proceed through trial and error. This setup requires the agent to build and adjust its mental model of the environment, recognizing key objects, spatial relationships, and potential interactions. Success depends on the agent's ability to hypothesize, test, and adapt its understanding. The task mirrors real-world reasoning challenges, where goals must be inferred, and solutions emerge through engagement.

\textbf{ ~2. Long-Horizontal successive task.}
In the absence of explicit guidance, the agent must achieve an average of \textbf{8.4 checkpoints} (critical tasks that act as bottlenecks in achieving goal completion), which requires an average of \textbf{27.2 steps} to successfully escape, even for the oracle trajectory. The considerable length of the task sequence and the dependencies between checkpoints require the agent to effectively utilize information from its past trajectory. Success hinges on the agent’s ability to integrate prior discoveries into its ongoing decision-making process.

\begin{figure*}[!htbp]
\centering
\begin{minipage}[c]{0.64\linewidth}
    \centering
    \includegraphics[width=\linewidth]{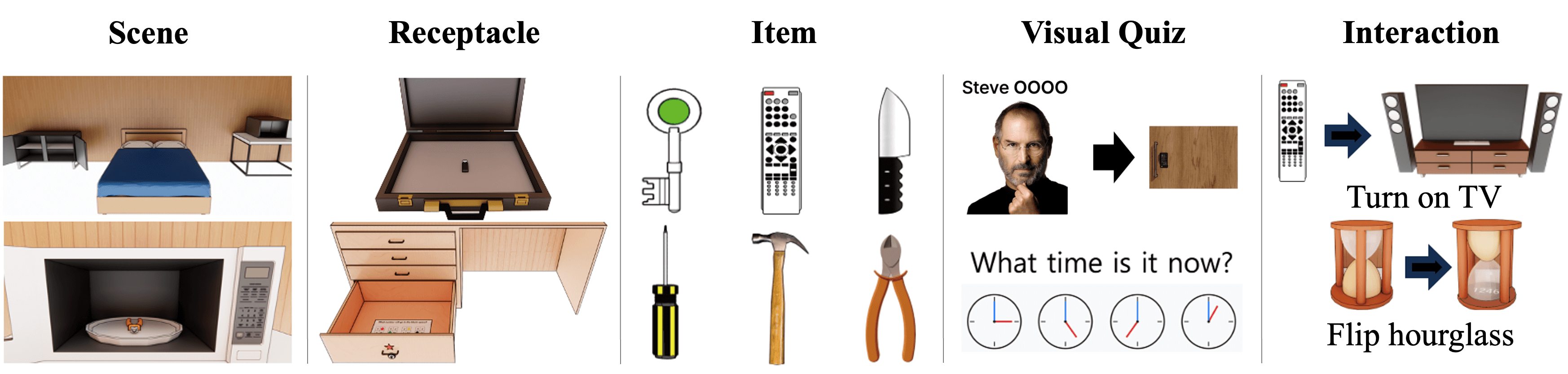}
    \caption{An illustration of each component in \datasetname.}
    \label{fig:example}
\end{minipage}\hfill
\begin{minipage}[c]{0.35\linewidth}
    \centering
    \scriptsize
    \setlength{\tabcolsep}{2pt}
    \begin{tabular}{c|ccccc}
    \toprule
    &Scene &Receptacle &Item &Quiz &Interaction \\
    \midrule
    \# Total &1084 &118 &102 &27 &311 \\
    \midrule
    \# Types &662 &32 &23 &27 &37 \\
    \midrule
    Avg./room &54.2 &5.9 &5.1 &1.4 &15.6 \\
    \bottomrule
    \end{tabular}
    \captionof{table}{Statistics for scenes, receptacles, items, quizzes, and interactions in \datasetname.}
    \label{tab:datastat_2}
\end{minipage}
\vspace{-5mm}

\end{figure*}

\section{Details on Dataset Composition and Construction}
\subsection{Dataset Composition}

\datasetname consists of 20 distinct rooms. Each room comprises four walls corresponding to the four cardinal directions, containing multiple objects classified into two types: \textit{Receptacles} and \textit{Items}, following the object categorization criteria used in~\cite{Shridhar_2020_CVPR}. \textit{Receptacles} are objects fixed within a room; they can be secured with locks or locking mechanisms and are capable of concealing or containing items. \textit{Items} can be collected into the inventory, and serve two functions: facilitating interactions with other receptacles or providing clues to unlock locks. \datasetname has 32 different receptacles and 23 different items, with each room containing a distinctive visual quiz. Each room contains an average of 5.9 receptacles and 5.1 items, along with one or two visual quizzes. Statistics and examples are presented in Table~\ref{tab:datastat_2} and Figure~\ref{fig:example}.

\textbf{~Visual Quizzes.}
Escape rooms challenge players to connect diverse pieces of information gathered through exploration, often by linking elements that initially appear unrelated. For example, players may need to associate similar symbols found in different locations or identify how objects of the same color are interconnected. Recognizing that escape rooms are well-suited for assessing the capability of AI models to draw inferences from environmental cues and formulate hypotheses for problem-solving, we incorporate visual quizzes. We integrate one to two number or letter code locks paired with corresponding hand-crafted visual quizzes into each room.

\textbf{~Game Graph.} To track the dynamically changing state of the room and provide corresponding observations, we use a Finite State Machine (FSM). FSM consists of two components: \textbf{states}, which correspond to the visual observations provided to the agent, and \textbf{transitions}, which define state changes based on player actions and specific conditions being met.

\textbf{~Actions.}
Agent can perform two types of actions: movement and object interaction. For movement, the agent can \textit{inspect [object]} to examine it closely or \textit{step back} to gain a broader view. Additionally, it can \textit{turn to [direction]} to view walls that are not currently visible. For object interaction, the agent can use acquired items (\eg Unlock with key) or perform actions associated with visible objects (\eg Turn on laptop). When the agent observes a lock requiring a quiz answer, an additional action to input answer becomes available. 

At each gamestep, the agent is given actions for movement and other possible actions based on its current state and observation. When the agent performs an action that satisfies an object's predefined conditions or corresponds to the lock's answer, it triggers state transitions according to game graph. See Figure~\ref{fig:wordcloud} for examples of object interactions.

\begin{figure}[!t]
    \centering
    \includegraphics[width=\linewidth]{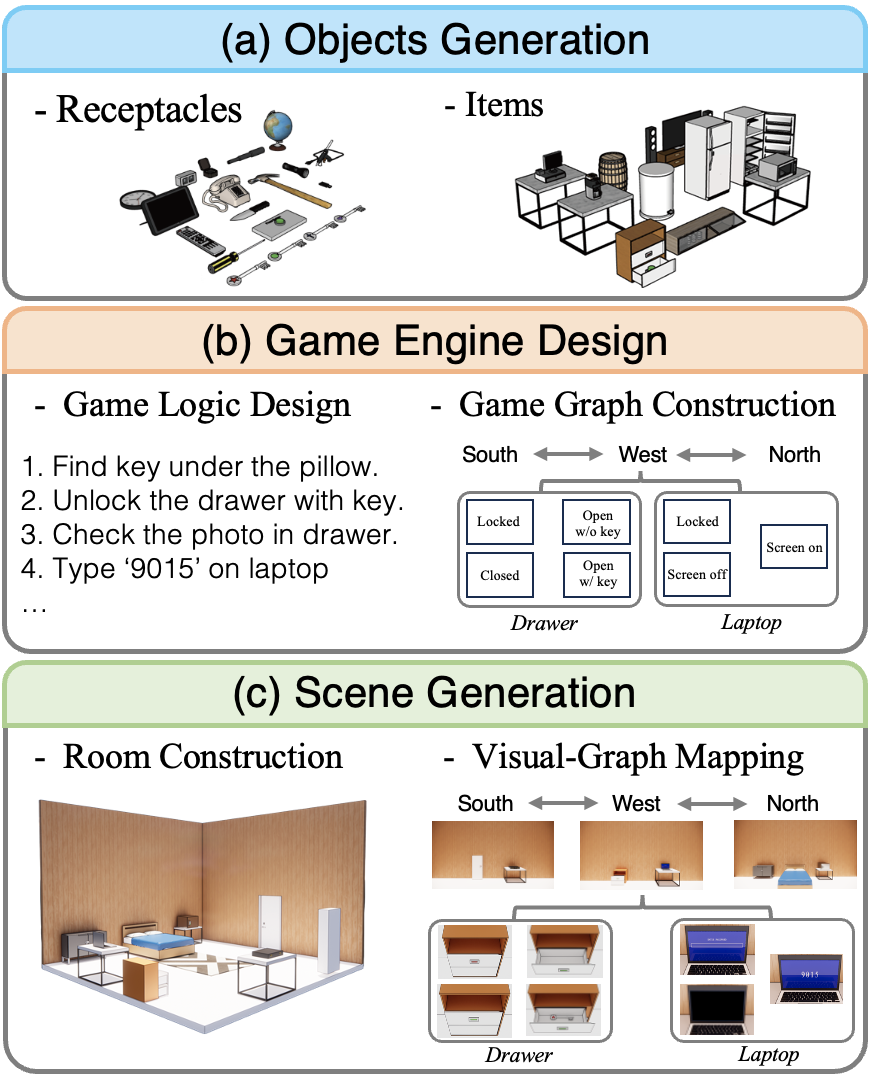}
    \caption{Process of \datasetname construction.}
    \label{fig:data_creation}
    \vspace{-5mm}
\end{figure}

\begin{table*}[h]
\centering

\resizebox{0.9\linewidth}{!}{%
\begin{tabular}{@{}l cccc l cccc@{}} %
\toprule
& \multicolumn{4}{c}{\textit{Frontier MLLMs}} & &\multicolumn{4}{c}{\textit{Open-source MLLMs}} \\
\cmidrule(lr){2-5} \cmidrule(lr){7-10}
\textbf{Model} & \textbf{SR$\uparrow$(\%)} & \textbf{GC$\uparrow$(\%)} & \textbf{SPL$\uparrow$(\%)} & \textbf{Step$\downarrow$} & \textbf{Model} & \textbf{SR$\uparrow$(\%)} & \textbf{GC$\uparrow$(\%)} & \textbf{SPL$\uparrow$(\%)} & \textbf{Step$\downarrow$} \\
\midrule

\claude & \textbf{6.3} & 11.11 & \textbf{3.35} & \textbf{286.7} & 

\qwenvl & 1.7 & 5.71 & 0.26 & 298.1 \\

\gpt & 3.3 & 16.00 & 0.60 & 294.5 &

\internfive & 0.0 & 5.75 & - & - \\

\gptmini& 0.0 & 5.37 & - & - &

\interntwo & 0.0 & 0.0 & - & - \\

\geminipro & 0.0 & \textbf{18.51} & - & - & 

\llava& 0.0 & 0.0 & - & - \\

\geminiflash & 0.0 & 9.54 & - & - & 

\llavaov & 0.0 & 0.0 & - & - \\
\midrule
\textbf{Human} & 82.5 & 95.80 & 51.30 & 52.8 \\
\bottomrule
\end{tabular}
}
\caption{Performance comparison of various MLLMs and human performance on the \datasetname benchmark. The best results for each metric are \textbf{bolded}. Refer Appendix~\ref{sec:human_study} for detailed information on the human study.  For models that failed to escape entirely, the standard deviation is marked with "-".}
\label{tab:baseline}
\vspace{-5mm}
\end{table*}

\subsection{\datasetname Construction}

 Figure~\ref{fig:data_creation} shows the creation process for rooms in \datasetname. \textbf{(a)} Objects are created using 3D modeling software, and those required for arrangement in each room were sampled. \textbf{(b)} Using the sampled objects, game logic that defines key checkpoints to complete the game is designed through AI-human collaboration. Game graph is then constructed as an FSM to track the overall game state and enable state transitions triggered by agent's actions. \textbf{(c)} Objects are arranged within the room according to its setup, and visual observations are mapped to corresponding game states.
See Appendix~\ref{sec:datasetdetail} for detailed explanations of each process and prompts used for game logic design.

\section{Examining MLLMs on \datasetname}
We evaluate the performance of various MLLMs, ranging from closed-source frontier models to open-source models on the \datasetname benchmark.
\subsection{Experiment Setup}
\label{sec:experiment_setup}
At each step, the model receives the current observation, history of the last 20 observations and actions, held items, and available actions, from which it selects one executable action. For each room, we conducted three different trials and averaged the resulting metrics. Each experiment was terminated early if no significant progress occurred within 100 consecutive steps and was automatically concluded after 300 steps. 

\subsection{Metrics}
We adopted widely used metrics for evaluating embodied agents. These include: \textbf{1. Success Rate~(SR)}, which is the ratio of escape success in an episode; \textbf{2. Goal Completion~(GC)}, the ratio of checkpoints completed at the end of an episode; \textbf{3. Step}, the total actions taken by an agent; and \textbf{4. Success weighted by Path Length~(SPL)}~\cite{anderson2018evaluation}, which penalizes success rate by total number of steps taken.

\subsection{Results}
Table~\ref{tab:baseline} presents the experimental results. For open-source models, none of the open-source models achieved a single success, except for \qwenvl. Notably, even sophisticated models such as \claude, \gpt, and \geminipro~ demonstrated success rates below 10\%. 
These results underscore the highly challenging nature of the exploration-centric problem-solving environment provided by \datasetname for AI models, concurrently highlighting significant avenues for improvement.

For models to perform well in \datasetname, they must effectively execute two primary capabilities that \datasetname requires: exploration and iterative hypothesis formulation and test. Rather than devising a specialized framework for the unique `escape room' task, we aim to analyze how modules considered crucial in agent modeling contribute to the exploration-centric problem-solving tasks provided by this benchmark. To this end, we augment \modelname with Memory management and Reasoning modules.

\begin{figure*}[!htbp]
\centering
\includegraphics[width=\linewidth]{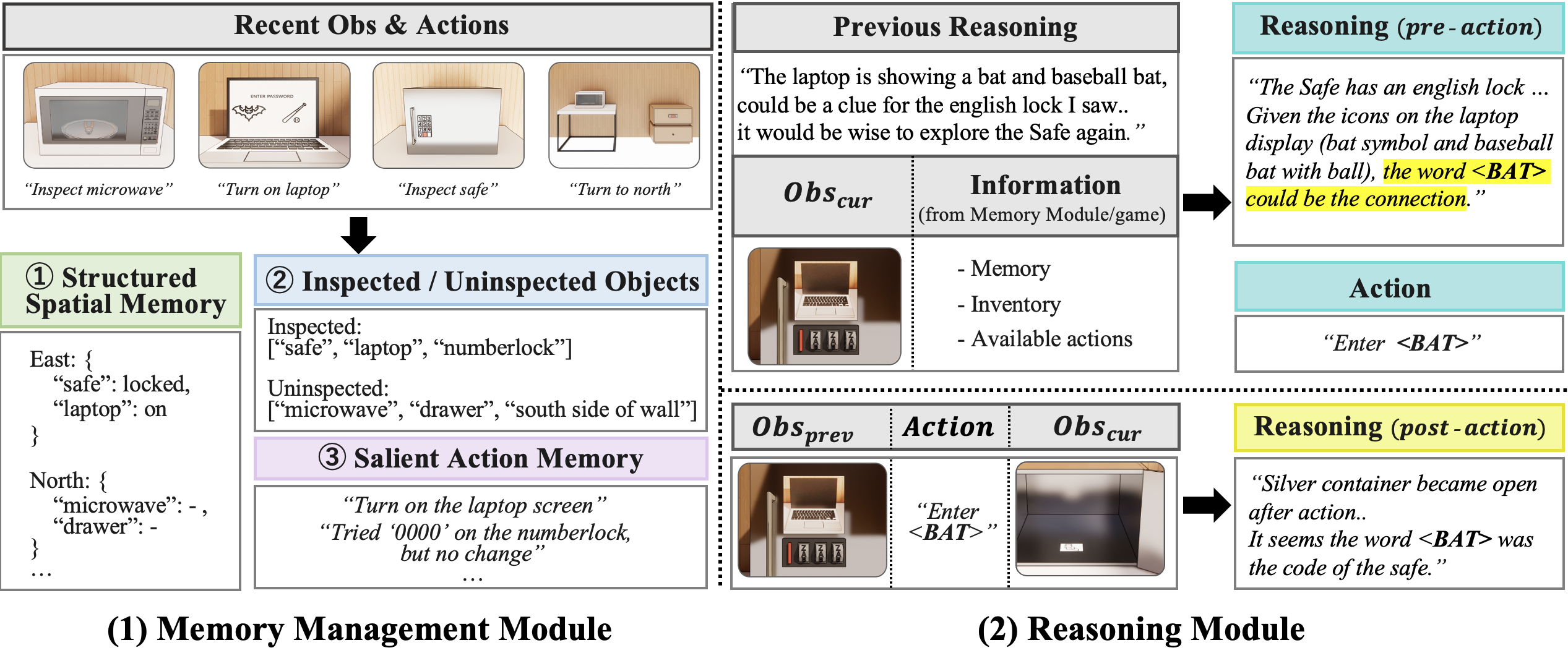}
\caption{An overview of memory management module and reasoning module, along with examples of inputs (gray boxes) and outputs (colored boxes) for each module.}
\label{fig:agent}
\vspace{-3mm}

\end{figure*}

\section{\modelname: Exploring Agent in \datasetname}

\subsection{Memory Management Module}
In \datasetname, agent can only observe a limited portion of the room. For effective navigation and exploration, the agent should maintain spatial memory that the agent constructs to form a coherent mental map of the space. Additionally, the agent must remember the actions it has attempted to learn from its trials and avoid redundant efforts. 

Therefore, we designed and applied a Memory Management module for the game agent~(Figure~\ref{fig:agent}-(1)). The memory module consists of \textbf{structured spatial memory}, \textbf{exploration memory}, and \textbf{salient action memory}, which are constructed and updated periodically. It compressively and structurally manages memory from observation and action histories accumulated over tens to hundreds of steps, while also updating periodically. This allows the agent to make clear decisions from a condensed history, effectively mitigating its processing overload by addressing both cognitive demands and the model's inherent context window constraints.

For the experiment, memory construction begins after the first 10 steps. Subsequently, memory management is performed every 10 steps. To avoid exceeding the context length and performance drop~\cite{zhao2024benchmarking} with multiple image inputs, we used image captions generated by the same model to replace visual observations with textual descriptions. Refer to Appendix~\ref{sec:memorymodule} for more details.

\subsection{Reasoning Module}
Exploration within escape rooms necessitates a cycle of hypothesis formulation and test. Specifically, the agent progresses by: 1) formulating hypotheses based on its memory and available information before action-decision, and 2) analyzing the outcomes to determine the validity of these hypotheses. Thus, we designed pre-action and post-action reasoning for hypothesis formulation and verification.

Pre-action reasoning, inspired by~\cite{yao2023react}, requires the agent to 'think' before action selection. This approach not only facilitates more effective action decision but also enables the decomposition and analysis of the model's reasoning process; we find that the model formulates hypotheses in various ways. We will discuss this in~\S\ref{sec:reasoning}.

Post-action reasoning involves the agent assessing the consequences of its actions, informed by the executed actions and the observations made immediately before and after those actions. This allows for hypothesis evaluation, and the results feed into the subsequent pre-action reasoning. The agent might then formulate different hypotheses or attempt new explorations, leading to more successful subsequent attempts.

\begin{table*}[h]
\centering
\resizebox{0.9\linewidth}{!}{%
\begin{tabular}{@{}l cccc@{}} %
\toprule
& \multicolumn{4}{c}{BaseAgent $ \rightarrow $ \modelname (\textit{diff.})} \\ %
\cmidrule(lr){2-5}
\textbf{Model} & \textbf{SR$\uparrow$(\%)} & \textbf{GC$\uparrow$(\%)} & \textbf{SPL$\uparrow$(\%)} & \textbf{Step$\downarrow$} \\
\midrule
\rowcolor{gray!20} %
\multicolumn{5}{l}{\textit{Frontier MLLMs}} \\ %
\claude & 6.7 $ \rightarrow $ 21.7~($\uparrow$15.0) & 11.11 $ \rightarrow $ 32.37 ~($\uparrow$21.26) & 3.35 $ \rightarrow $ \textbf{10.92} ~\underline{($\uparrow$7.57)} & 286.7 $ \rightarrow $ \textbf{249.5} ~($\downarrow$37.2) \\

\gpt & 3.3 $ \rightarrow $ 13.3 ~~($\uparrow$10.0) & 16.00 $ \rightarrow $ 29.02 ~($\uparrow$13.02) & 0.60 $ \rightarrow $ 3.91 ~($\uparrow$3.31) & 294.5 $ \rightarrow $ 276.2 ~($\downarrow$18.3) \\

\gptmini & 0.0 $ \rightarrow $ 6.7 ~($\uparrow$6.7) & 5.37 $ \rightarrow $ 21.31 ~($\uparrow$15.94) & 0.00 $ \rightarrow $ 1.05 ~($\uparrow$1.05) & 300.0 $ \rightarrow $ 291.2 ~($\downarrow$8.8) \\

\geminipro & 0.0 $ \rightarrow $ \textbf{25.0} ~\underline{($\uparrow$25.0)} & 18.51 $ \rightarrow $ \textbf{42.16} ~($\uparrow$23.65) & 0.00 $ \rightarrow $ 6.59 ~($\uparrow$6.59) & 300.0 $ \rightarrow $ 258.1 ~($\downarrow$41.9) \\

\geminiflash & 0.0 $ \rightarrow $ 23.3 ~($\uparrow$23.3) & 9.54 $ \rightarrow $ 34.40 ~\underline{($\uparrow$24.86)} & 0.00 $ \rightarrow $ 6.32 ~($\uparrow$6.32) & 300.0 $ \rightarrow $ 255.0 ~\underline{($\downarrow$45.0)} \\

\midrule
\rowcolor{gray!20}

\multicolumn{5}{l}{\textit{Open-source MLLMs}} \\ %
\qwenvl & 1.7 $ \rightarrow $ 3.3 ~($\uparrow$1.6) & 5.71 $ \rightarrow $ 11.46 ~($\uparrow$5.75) & 0.26 $ \rightarrow $ 0.42 ~($\uparrow$0.16) & 298.1 $ \rightarrow $ 295.8 ~($\downarrow$2.3) \\

\internfive & 0.0 $ \rightarrow $ 0.0 (-) & 5.75 $ \rightarrow $ 6.20 ~($\uparrow$0.45) & 0.00 $ \rightarrow $ 0.00 (-) & 300.0 $ \rightarrow $ 300.0 (-) \\

\interntwo & 0.0 $ \rightarrow $ 0.0 (-) & 2.88 $ \rightarrow $ 4.91 ~($\uparrow$2.03) & 0.00 $ \rightarrow $ 0.00 (-) & 300.0 $ \rightarrow $ 300.0 (-) \\

\llava & 0.0 $ \rightarrow $ 0.0 (-) & 0.17 $ \rightarrow $ 0.54 ($\uparrow$0.37) & 0.00 $ \rightarrow $ 0.00 (-) & 300.0 $ \rightarrow $ 300.0 (-) \\
\llavaov & 0.0 $ \rightarrow $ 0.0 (-) & 1.08 $ \rightarrow $ 1.25 ~($\uparrow$0.17) & 0.00 $ \rightarrow $ 0.00 (-) & 300.0 $ \rightarrow $ 300.0 (-) \\
\bottomrule
\end{tabular}
}
\caption{Comparison of performance before and after applying the memory management and reasoning modules. For ease of reference, the agent operating without these two modules (Table~\ref{tab:baseline}) is referred to as `BaseAgent'. Standard deviation for steps is in Table~\ref{tab:step_std}.}
\label{tab:baseline_to_ours}

\end{table*}

\subsection{Results}

Table~\ref{tab:baseline_to_ours} shows scores of \modelname and the difference compared to BaseAgent. Frontier models improved across all metrics. For open-source models, goal completion generally increased, though eventually not enough for room escape. However, the performance of \qwenvl, driven by the adoption of memory management and reasoning, increased to a level comparable to \claude. Notably, although higher success rates and goal completion typically require more steps, all models with \modelname nevertheless demonstrated a reduction in trajectory length, leading to an improvement in SPL. These results underscore that agents become more effective (achieving more) and more efficient (within fewer steps) when they are enabled with memory management and reasoning. 

\begin{figure}[t]
\centering
\includegraphics[width=\linewidth]{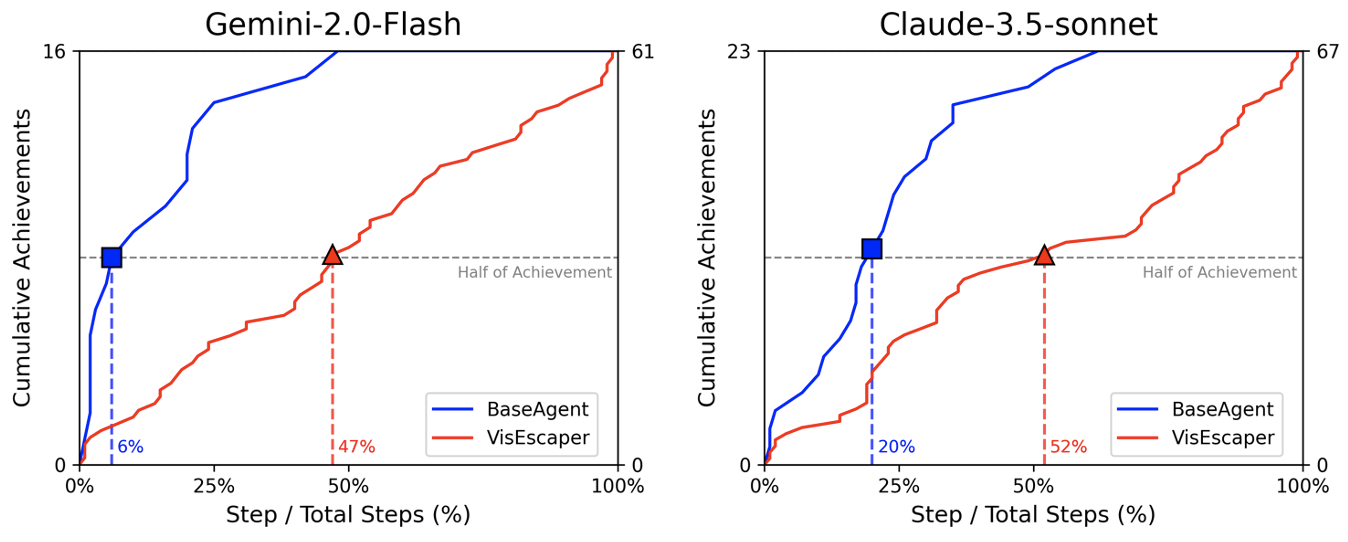}
\caption{Cumulative goal completion over trajectory time steps for BaseAgent and \modelname in each room. For comparison, each agent's progression is shown on the same scale. Additionally, the point where 50\% of goal completion is achieved is marked.}
\label{fig:success_timing}
\vspace{-5mm}

\end{figure}

Furthermore, the design of \modelname enables the agent to achieve steady progress without becoming disoriented within the expanding information space. Figure~\ref{fig:success_timing} illustrates the cumulative goal completion for BaseAgent and \modelname over the course of a trajectory. Progress by BaseAgent is concentrated early in the trajectory (half of its achievements are realized within the initial 6\% of steps for \geminiflash~and 20\% for \claude).
In contrast, \modelname demonstrates consistent progression.

Upon analyzing model trajectories, we observed failures predominantly occur in visual quizzes and are more pronounced in open-source models, which achieved smaller performance gains than frontier models even with memory and reasoning modules. We identified two primary reasons for this gap.

First, reasoning modules significantly boost visual quiz accuracy for Frontier models, an effect less evident in open-source models (Appendix~\ref{sec:reason_quiz}). Second, compared to Frontier models, open-source models tend to explore evidence less thoroughly and more frequently resort to brute-force attempts-resulting in meaningless and repetitive actions, ultimately causing them to get stuck in the game (Appendix~\ref{sec:brute-force}).

\begin{table}[h]
\centering
\resizebox{\linewidth}{!}{
\begin{tabular}{lccccc}

\toprule
&  \multicolumn{5}{c}{\textbf{\exphint}} \\
\cmidrule(lr){2-6}
\textbf{Model}& \textbf{SR(\%)} & \textbf{GC(\%)} & \textbf{HCR$\downarrow$(\%)} & \textbf{SPL  (\%)} & \textbf{Step} \\
\midrule

\rowcolor{gray!20} %
\multicolumn{6}{l}{\textit{Frontier MLLMs}} \\

\claude & 90.0 & 96.33 & \textbf{18.87} & \textbf{30.46} & \textbf{103.1} \\
\gpt & \textbf{95.0} & \textbf{97.39} & 20.74 & 26.83 & 112.9 \\
\gptmini & 46.7 & 67.40 & 32.16 & 8.93 & 228.2 \\
\geminipro & 71.7&82.53&20.18&18.22&163.3 \\
\geminiflash & 66.7 & 75.78 & 25.11 & 15.56 & 179.2\\

\rowcolor{gray!20} %
\multicolumn{6}{l}{\textit{Open-source MLLMs}} \\
\qwenvl & 61.7 & 73.13 & 30.51 & 12.07 & 197.0 \\
\internfive & 66.7 & 79.23 & 41.69 & 12.23 & 210.5 \\
\interntwo  & 8.3 & 48.18 & 61.61 & 0.96 & 291.1 \\
\llava & 15.0 & 34.79 & 64.97 & 1.47 & 289.8 \\
\llavaov & 8.3 & 41.02 & 62.89 & 0.80 & 294.9 \\

\midrule
\end{tabular}
}
\caption{Results when hints are allowed for models. \textbf{HCR} means \textbf{Hint-assisted Completion Rate}: the ratio of checkpoints completed by the agent with the assistance of hints, where lower is better.}
\label{tab:main_hint}
\end{table}

\subsection{Results on Hint-guided Experiment}
However, despite overall improvements, if the model becomes deadlocked in certain challenging checkpoints and thereby blocks subsequent progress, comprehensive experimental analysis can become difficult. Thus, we also experimented with hint-guided setting; if the model is stuck for 30 steps or more, a guideline message for the next checkpoint is provided to the model. More details on hint-guided experiment are in Appendix~\ref{sec:hint}.

A comparison of performance between scores from \modelname of Table~\ref{tab:baseline_to_ours} and Table~\ref{tab:main_hint} shows that after hints are allowed, all models demonstrated a significant increase in success rate and goal completion. Notably, \claude~and \gpt~escaped rooms using fewer than half the steps compared to before receiving hints, leading to the greatest improvement, especially in SPL, which captures performance and efficiency together. Also, smaller models still improve substantially with hints, but rely on hints more frequently, as shown by higher HCR.

\section{Deep Analysis on Each Module}
\subsection{Memory Management}
\begin{table}[!htp]\centering

\scriptsize
\resizebox{0.95\linewidth}{!}{

\begin{tabular}{lcc|cc}\toprule
Model & \textbf{\textit{None.}} &\textbf{\textit{Me.}} &\textbf{\textit{Re.}} &\textbf{\textit{Me}\&\textit{Re}} \\\midrule

\claude &29.1 &\textbf{24.3} &8.5 &\textbf{5.7} \\
\gpt &27.8 &\textbf{26.1} &20.3 &\textbf{7.9} \\
\internfive &27.5 & 30.2 &34.8 &\textbf{21.6} \\
\llava &53.7 &\textbf{50.8} &34.1 &\textbf{24.6} \\
\midrule
\textit{Diff. (Avg)} & \multicolumn{2}{c}{-1.6} & \multicolumn{2}{c}{\textbf{-9.5}} \\ 
\bottomrule
\end{tabular}
}
\caption{Ratio of repetitive actions. \textbf{\textit{None}}-BaseAgent, \textbf{\textit{Me.}}-BaseAgent +memory, \textbf{\textit{Re.}}-BaseAgent +reasoning, and \textbf{\textit{Re.\&Me.}}-\modelname. \textit{Diff (Avg).} denotes the averaged difference with or without memory.}
\label{tab:dup_base_vs_ours}
\end{table}

\textbf{~Repetitive Actions.} Ideally, an escape room player with good memory is less likely to repeat actions or re-explore areas. Consequently, to investigate how memory management influences an agent's action decision within a trajectory, we examined the action repetition in a trajectory from diverse models. We only included actions related to object interaction and answer submission, since \textit{inspect, turn, step back} are part of the exploration process. 

According to Table~\ref{tab:dup_base_vs_ours}, memory management enabled efficient action decision by reducing the overall rate of repetition actions. However, the effect was greater when memory management was performed with reasoning (-9.5\%) compared to when it was performed without reasoning (-1.6\%). This suggests that the synergy between memory management and reasoning capabilities allows the agent to more effectively utilize past experiences to avoid redundant actions and make more optimal decisions. Analysis of difference between models are suggested in Appendix~\ref{sec:repetition}.

\subsection{Reasoning}
\label{sec:reasoning}
\begin{figure}[!t]
    \centering
    \includegraphics[width=\linewidth]{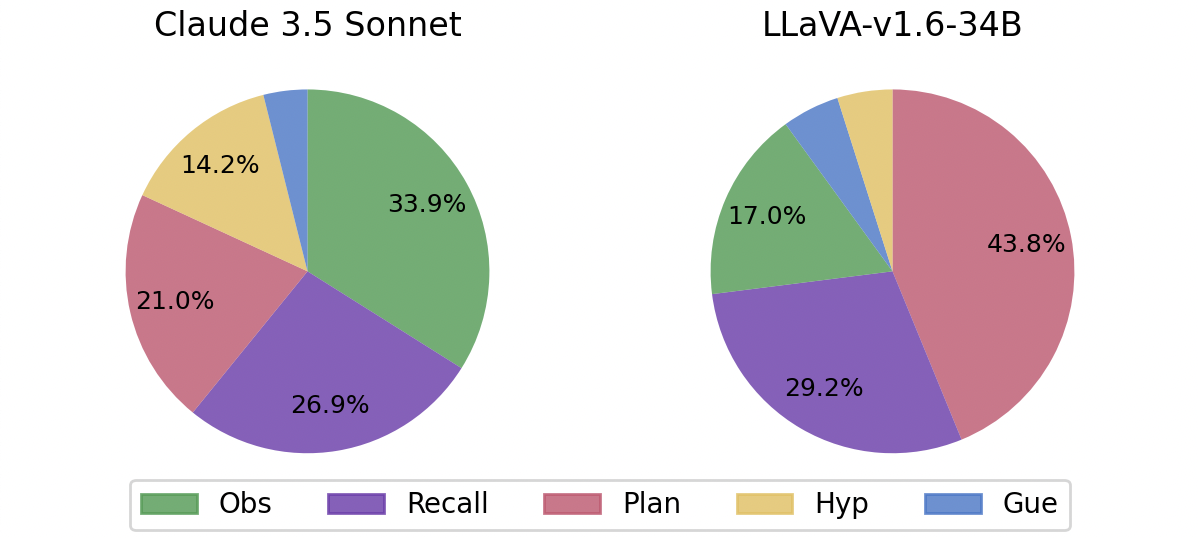}
    \caption{Proportion of each reasoning component.}
    \label{fig:reasoning_pie}
    \vspace{-8mm}
\end{figure}

Adopting reasoning process before action decision enables analyzing how agents in \datasetname perform reasoning and how they differ across models. By dividing their reasoning sentence into five components: Observation, Recall, Planning, Hypothesizing and Guess, we analyzed how they utilize these components and combine these components to perform reasoning. For categorization, we broke down the reasoning process and performed a machine-based evaluation for each piece with GPT-4o ~\cite{achiam2023gpt}. Specifically, we analyzed two models (\claude~and \llava). The proportion of each component is shown in Figure~\ref{fig:reasoning_pie}, and its composition is in Table~\ref{tab:category_example}. The top 3 most frequent component combinations were reported. See Appendix~\ref{sec:reasoning_component} for the criteria used to divide each module and human assessment for ensuring reliability.

Through decomposition, we find following results: \textbf{(Fig.~\ref{fig:reasoning_pie}) Memory Recall.} Across both models, they frequently referenced memory~(26.9\% and 29.2\%). This highlights that reasoning can be influenced by the establishment of good memory. \textbf{(Tab.~\ref{tab:category_example}) Reasoning Composition.} \claude~exhibits longer reasoning sequences, systematically combining various components. Notably, their reasoning often proceeds after revisiting current observations and past memories. In contrast, \llava~demonstrates simpler and shorter reasoning sequences. \textbf{(Fig.~\ref{fig:reasoning_pie} \&Tab.~\ref{tab:category_example}) Hypothesis Formulation.} \claude~frequently attempts to formulate reasoned hypotheses, often preceded by analyzing the current observation and consulting its memory before formulating hypothesis, while \llava~rarely does so.

\begin{table*}[!htp]\centering
\scriptsize
\resizebox{\textwidth}{!}{
\begin{tabular}{llccc}
\toprule
\textbf{Model} & \textbf{Composition} &  \textbf{Ratio} & \textbf{Length} & \textbf{Example}  \\
\midrule
\multirow{3}{*}[-4mm]{\claude} 
  & \textit{(\textbf{\textcolor{red}{Obs}}, \textcolor{blue}{Recall}, \textcolor{brown}{Plan})}
  & \gradientcell{18.1} & 127.0 
  & \parbox[c]{.7\textwidth}{
      \centering  
      \textcolor{red}{\textbf{\textbf{[\textit{Obs}]}}} I'm facing Door with Keycard lock.
      \textcolor{blue}{\textbf{\textbf{[\textit{Recall}]}}} I haven't specifically tried Entrancecard. \\
      \textcolor{brown}{\textbf{\textbf{[\textit{Plan}]}}} I should inspect the lock.
    } \\
\cmidrule(lr){5-5}  
& \textit{(\textbf{\textcolor{red}{Obs}}, \textcolor{blue}{Recall}, \textbf{\textcolor{teal}{Hypo}}, \textcolor{brown}{Plan})}   
  & \gradientcell{15.8} & 141.5 
  & \parbox[c]{.7\textwidth}{
      \centering  
      \textcolor{red}{\textbf{[\textit{Obs}]}} We're facing a door lock.
      \textcolor{blue}{\textbf{[\textit{Recall}]}} Seems to connect with the kiosk's display. \\
      \textcolor{teal}{\textbf{[\textit{Hypo}]}} This suggests we need to find four words.
      \textcolor{brown}{\textbf{[\textit{Plan}]}} Return to the kiosk to check.
    } \\
\cmidrule(lr){5-5}
& \textit{(\textbf{\textcolor{red}{Obs}}, \textcolor{blue}{Recall}, \textcolor{brown}{Plan}, \textbf{\textcolor{teal}{Hypo}})}   
  & \gradientcell{14.8} & 127.5 
  & \parbox[c]{.7\textwidth}{
      \centering  
      \textcolor{red}{\textbf{[\textit{Obs}]}} We're currently facing locked box.
      \textcolor{blue}{\textbf{[\textit{Recall}]}} Already tried some number combinations. \\
      \textcolor{brown}{\textbf{[\textit{Plan}]}} Re-checking east wall poster might be worthwhile.
      \textcolor{teal}{\textbf{[\textit{Hypo}]}} We may find all primes numbers in poster.
    } \\
\midrule
\multirow{3}{*}[-3mm]{\llava}
  & \textit{(\textcolor{blue}{Recall}, \textcolor{brown}{Plan})}
  & \gradientcell{38.5} & 55.3 
  & \parbox[c]{.7\textwidth}{
    \centering
    \textcolor{blue}{\textbf{[\textit{Recall}]}} I've already inspected wardrobe \&door. 
    \textcolor{brown}{\textbf{[\textit{Plan}]}} I should inspect sensor lock on the door.
  }\\
\cmidrule(lr){5-5}
& \textit{(\textcolor{brown}{Plan})}
  & \gradientcell{21.3} & 39.4 
  & \parbox[c]{.7\textwidth}{
    \centering
    \textcolor{brown}{\textbf{[\textit{Plan}]}} I will inspect coffee machine to see if there are any clues or items.
  }\\
\cmidrule(lr){5-5}
& \textit{(\textcolor{red}{Obs}, \textcolor{brown}{Plan})}
  & \gradientcell{11.7} & 54.4  
  & \parbox[c]{.7\textwidth}{
    \centering
    \textcolor{red}{\textbf{[\textit{Obs}]}} Locker is currently locked.   \textcolor{brown}{\textbf{[\textit{Plan}]}} I should try to unlock it to see if there are any clues.
  }\\
\bottomrule
\end{tabular}
}
\caption{The top 3 compositions of reasoning components most frequently utilized by \claude~and \llava. Token lengths are calculated using the `token-count' in pypl, setting model as gpt-3.5-turbo.}
\label{tab:category_example}
\end{table*}

\begin{table*}[!htp]
\centering
\resizebox{\textwidth}{!}{
    \scriptsize
    \begin{tabular}{cccccccccccc}
    \toprule
    & & \multicolumn{4}{c}{\textbf{\expbase}} & \multicolumn{5}{c}{\textbf{\exphint}} \\
    \cmidrule(lr){3-6}\cmidrule(lr){7-11}
    
    \textbf{Model(MLLM / LLM)} & \textbf{Input Modality} & \textbf{SR$\uparrow$(\%)} & \textbf{GC$\uparrow$(\%)} & \textbf{SPL$\uparrow$(\%)} & \textbf{Step$\downarrow$} & \textbf{SR$\uparrow$(\%)} & \textbf{GC$\uparrow$(\%)} & \textbf{HCR$\downarrow$(\%)} & \textbf{SPL$\uparrow$(\%)} & \textbf{Step$\downarrow$} \\
    \midrule

    InternVL2.5-38B & Image &0.0 &6.20 &- &- &66.7 &79.23 &41.69 &12.23 &210.5 \\
    InternVL2.5-38B + Base LLM & Caption &1.7 &9.57 &0.67 &295.9 &35.0 &53.76 &38.14 &7.08 &245.3 \\
    InternVL2.5-38B + Base LLM~(R1) & Caption &\textbf{10.3} &\textbf{19.58} &\textbf{7.81} &\textbf{261.6} &\textbf{80.0} &\textbf{87.06} &\textbf{25.53} &\textbf{19.37} &\textbf{166.8} \\
    
    \midrule
    InternVL2-40B & Image &0.0 &4.91 &- & - &8.3 &48.18 &61.61 &0.96 &291.1 \\
    InternVL2-40B + Base LLM & Caption &0.0 &3.88 &- &- &\textbf{32.0} &\textbf{68.00} &\textbf{48.22} &\textbf{4.39} &\textbf{250.6} \\
    \midrule
    LLaVA-v1.6-34B & Image &0.0 &0.54 &- &- &15.0 &34.79 &64.97 &1.47 &289.8 \\
    LLaVA-v1.6-34B + Base LLM & Caption &0.0 &\textbf{4.14} &- &- &\textbf{38.3} &\textbf{61.80} &\textbf{50.08} &\textbf{4.59} &\textbf{257.7} \\
    \midrule
    LLaVA-OneVision-7B & Image &0.0 &1.25 &- &- &8.3 &41.02 &62.89 &0.80 &294.9 \\
    LLaVA-OneVision-7B + Base LLM & Caption &0.0 &\textbf{1.62} &- &- &\textbf{25.0} &\textbf{58.93} &70.70 &\textbf{2.40} &\textbf{287.3} \\

    \bottomrule
    \end{tabular}
}
\caption{Comparative performance of results from MLLMs and Socratic method. Experiment is conducted by models with memory management and reasoning module. \expbase~and \exphint ~denote experiments conducted without hints and hint-guided experiments, respectively.}
\label{tab:socratic}
\vspace{-5mm}
\end{table*}

\section{Replacing VLM with LLM}
Socratic models, as represented by \cite{shin2024socratic}, is an approach that uses image captioners to convert visual inputs into descriptive text inputs and then uses LLMs to perform multimodal inference. Given that MLLMs excel at visual perception but often struggle with complex visual reasoning~\cite{li2024enhancing, gan2022vision, liu2023matcr}, we adopt LLMs to address this gap. 

We compared the results from two experimental setups as \S\ref{sec:experiment_setup}: one where the MLLM directly processed the visual inputs, and another where a LLM processed image captions generated by an MLLM. 
For controlled experiments, we used only the backbone LLM of the corresponding MLLM-refer to Table~\ref{tab:corresponding_llm}. Across the models investigated, excluding \internfive, improvements were observed in most metrics when socratic method was adopted instead of MLLMs.

Notably, for \internfive, which underperformed with \textit{Qwen-32B-Instruct} with hints, switching to \textit{Deepseek-R1-Distill-Qwen-32B}-sharing the same architecture but with reasoning capabilities enhanced by distillation from Deepseek-R1~\cite{guo2025deepseek}—markedly improved all evaluation metrics. These results suggest that: 1. Many MLLMs struggle with complex reasoning in multimodal contexts—a critical capability for \datasetname, and 2. Integrating strong reasoning capabilities of LLMs with visual perception of VLMs can lead to enhanced multimodal reasoning capability.

\section{Related Works}
\label{sec:relatedworks}
\subsection{Interactive Environments for Evaluating LLM/VLM}

Text-based games have long been used to benchmark LLMs, offering interactive settings to assess reasoning and behavior~\cite{cote2019textworld, hausknecht2020interactive, qian2024escapebench}. More recent benchmarks extend this to testing scientific knowledge~\cite{wang2022scienceworld} and ethical reasoning~\cite{pan2023rewards}. Multimodal environments like Minecraft and embodied agent simulators have further enabled evaluation on both LLM and VLM agents in visually grounded settings~\cite{wangvoyager, Odyssey2024, dong2024villageragent, guss2019minerl, ai2thor, puig2018virtualhome, shridharalfworld, habitat19iccv}. GUI-based benchmarks~\cite{deng2024multi, deng2023mind2web, xu2021grounding} now assess how these models perceive and act on web interfaces, testing their ability to recognize visual elements, reason about functions~\cite{shahbandeh2024naviqate}, and perform tasks~\cite{shi2017world, furuta2023exposing, yao2022webshop, zhouwebarena}.

\subsection{LLM/VLM as an Action Decision Maker}

The effective deployment of LLM/VLM agents in real-world applications requires adaptability to dynamic and unpredictable environments~\cite{mnih2013playing, nagabandi2019learning, zhang2018adaptive}. Agents must reason adaptively and learn continuously~\cite{zhou2024symbolic}.
This is support by three core capabilities: experience accumulation~\cite{feng2024agile} leveraging past interactions, strategic exploration~\cite{zhu2023ghost} balancing risk and learning, and knowledge abstraction~\cite{zhao2023expel, paglieri2024balrog} forming high-level representations.
Our agent incorporates two modules-Memory and Reasoning~\cite{yao2023react, ke2024hydra, zheng2025lifelong}-to support these core capabilities and enable effective performance in dynamic settings.

\section{Conclusion}
In this study, we introduce \datasetname, a novel benchmark of escape room scenarios designed to rigorously test multimodal agents' abilities to explore, reason, and make decisions in interactive, dynamic environments. We show that existing multi-modal models struggle under these challenging conditions. 
By integrating memory management and reasoning, we observed that agents could achieve more efficient exploration and reasoning, and noted the crucial synergy between memory and reasoning. Our findings underscore the importance of memory and adaptive reasoning for building more capable, autonomous AI systems in complex and real-world tasks.

\section{Limitations}
While \datasetname presents a valuable benchmark for testing the exploration-driven problem-solving abilities of multimodal agents, the specific task of room escapes may not fully generalize to real-world environments where agents encounter more complex situations and more unfamiliar objects. Additionally, since the action space of \datasetname is discrete rather than continuous, there is a need to convert the conclusions drawn from multiple modules into feasible continuous action sequences for real-world execution.

\bibliography{custom}

\begin{thebibliography}{55}
\providecommand{\natexlab}[1]{#1}

\bibitem[{Achiam et~al.(2023)Achiam, Adler, Agarwal, Ahmad, Akkaya, Aleman, Almeida, Altenschmidt, Altman, Anadkat et~al.}]{achiam2023gpt}
Josh Achiam, Steven Adler, Sandhini Agarwal, Lama Ahmad, Ilge Akkaya, Florencia~Leoni Aleman, Diogo Almeida, Janko Altenschmidt, Sam Altman, Shyamal Anadkat, and 1 others. 2023.
\newblock Gpt-4 technical report.
\newblock \emph{arXiv preprint arXiv:2303.08774}.

\bibitem[{Anderson et~al.(2018)Anderson, Chang, Chaplot, Dosovitskiy, Gupta, Koltun, Kosecka, Malik, Mottaghi, Savva et~al.}]{anderson2018evaluation}
Peter Anderson, Angel Chang, Devendra~Singh Chaplot, Alexey Dosovitskiy, Saurabh Gupta, Vladlen Koltun, Jana Kosecka, Jitendra Malik, Roozbeh Mottaghi, Manolis Savva, and 1 others. 2018.
\newblock On evaluation of embodied navigation agents.
\newblock \emph{arXiv preprint arXiv:1807.06757}.

\bibitem[{Anthropic()}]{claude3}
Anthropic.
\newblock \href {https://api.semanticscholar.org/CorpusID:268232499} {The claude 3 model family: Opus, sonnet, haiku}.

\bibitem[{Bai et~al.(2025)Bai, Chen, Liu, Wang, Ge, Song, Dang, Wang, Wang, Tang et~al.}]{bai2025qwen2_5vl}
Shuai Bai, Keqin Chen, Xuejing Liu, Jialin Wang, Wenbin Ge, Sibo Song, Kai Dang, Peng Wang, Shijie Wang, Jun Tang, and 1 others. 2025.
\newblock Qwen2. 5-vl technical report.
\newblock \emph{arXiv preprint arXiv:2502.13923}.

\bibitem[{Chen et~al.(2024)Chen, Wang, Cao, Liu, Gao, Cui, Zhu, Ye, Tian, Liu et~al.}]{chen2024expanding}
Zhe Chen, Weiyun Wang, Yue Cao, Yangzhou Liu, Zhangwei Gao, Erfei Cui, Jinguo Zhu, Shenglong Ye, Hao Tian, Zhaoyang Liu, and 1 others. 2024.
\newblock Expanding performance boundaries of open-source multimodal models with model, data, and test-time scaling.
\newblock \emph{arXiv preprint arXiv:2412.05271}.

\bibitem[{C{\^o}t{\'e} et~al.(2019)C{\^o}t{\'e}, K{\'a}d{\'a}r, Yuan, Kybartas, Barnes, Fine, Moore, Hausknecht, El~Asri, Adada et~al.}]{cote2019textworld}
Marc-Alexandre C{\^o}t{\'e}, Akos K{\'a}d{\'a}r, Xingdi Yuan, Ben Kybartas, Tavian Barnes, Emery Fine, James Moore, Matthew Hausknecht, Layla El~Asri, Mahmoud Adada, and 1 others. 2019.
\newblock Textworld: A learning environment for text-based games.
\newblock In \emph{Computer Games: 7th Workshop, CGW 2018, Held in Conjunction with the 27th International Conference on Artificial Intelligence, IJCAI 2018, Stockholm, Sweden, July 13, 2018, Revised Selected Papers 7}, pages 41--75. Springer.

\bibitem[{Deng et~al.(2023)Deng, Gu, Zheng, Chen, Stevens, Wang, Sun, and Su}]{deng2023mind2web}
Xiang Deng, Yu~Gu, Boyuan Zheng, Shijie Chen, Sam Stevens, Boshi Wang, Huan Sun, and Yu~Su. 2023.
\newblock Mind2web: Towards a generalist agent for the web.
\newblock \emph{Advances in Neural Information Processing Systems}, 36:28091--28114.

\bibitem[{Deng et~al.(2024)Deng, Zhang, Zhang, Yuan, Ng, and Chua}]{deng2024multi}
Yang Deng, Xuan Zhang, Wenxuan Zhang, Yifei Yuan, See-Kiong Ng, and Tat-Seng Chua. 2024.
\newblock On the multi-turn instruction following for conversational web agents.
\newblock \emph{arXiv preprint arXiv:2402.15057}.

\bibitem[{Dong et~al.(2024)Dong, Zhu, Pan, Zhu, and Yang}]{dong2024villageragent}
Yubo Dong, Xukun Zhu, Zhengzhe Pan, Linchao Zhu, and Yi~Yang. 2024.
\newblock \href {https://arxiv.org/abs/2406.05720} {Villageragent: A graph-based multi-agent framework for coordinating complex task dependencies in minecraft}.
\newblock In \emph{Proceedings of the 62nd Annual Meeting of the Association for Computational Linguistics (ACL)}.

\bibitem[{Feng et~al.(2024)Feng, He, Huang, Lin, Zhang, Zhang, and Li}]{feng2024agile}
Peiyuan Feng, Yichen He, Guanhua Huang, Yuan Lin, Hanchong Zhang, Yuchen Zhang, and Hang Li. 2024.
\newblock \href {https://arxiv.org/abs/2405.14751} {{AGILE}: A novel reinforcement learning framework of {LLM} agents}.
\newblock In \emph{Advances in Neural Information Processing Systems (NeurIPS)}.

\bibitem[{Furuta et~al.(2023)Furuta, Matsuo, Faust, and Gur}]{furuta2023exposing}
Hiroki Furuta, Yutaka Matsuo, Aleksandra Faust, and Izzeddin Gur. 2023.
\newblock Exposing limitations of language model agents in sequential-task compositions on the web.
\newblock \emph{arXiv preprint arXiv:2311.18751}.

\bibitem[{Gan et~al.(2022)Gan, Li, Li, Wang, Liu, and Gao}]{gan2022vision}
Zhe Gan, Linjie Li, Chunyuan Li, Lijuan Wang, Zicheng Liu, and Jianfeng Gao. 2022.
\newblock \href {https://arxiv.org/abs/2210.09263} {Vision-language pre-training: Basics, recent advances, and future trends}.
\newblock \emph{arXiv preprint arXiv:2210.09263}.

\bibitem[{{Google DeepMind}(2024)}]{google2024gemini}
{Google DeepMind}. 2024.
\newblock \href {https://blog.google/technology/google-deepmind/google-gemini-ai-update-december-2024/#ceo-message} {Google gemini ai update: December 2024}.
\newblock Accessed: 2025-05-20.

\bibitem[{Guo et~al.(2025)Guo, Yang, Zhang, Song, Zhang, Xu, Zhu, Ma, Wang, Bi et~al.}]{guo2025deepseek}
Daya Guo, Dejian Yang, Haowei Zhang, Junxiao Song, Ruoyu Zhang, Runxin Xu, Qihao Zhu, Shirong Ma, Peiyi Wang, Xiao Bi, and 1 others. 2025.
\newblock Deepseek-r1: Incentivizing reasoning capability in llms via reinforcement learning.
\newblock \emph{arXiv preprint arXiv:2501.12948}.

\bibitem[{Guss et~al.(2019)Guss, Houghton, Topin, Wang, Codel, Veloso, and Salakhutdinov}]{guss2019minerl}
William~H Guss, Brandon Houghton, Nicholay Topin, Phillip Wang, Cayden~R Codel, Manuela Veloso, and Ruslan Salakhutdinov. 2019.
\newblock Minerl: A large-scale dataset of minecraft demonstrations.
\newblock \emph{arXiv preprint arXiv:1907.13440}.

\bibitem[{Hausknecht et~al.(2020)Hausknecht, Ammanabrolu, C{\^o}t{\'e}, and Yuan}]{hausknecht2020interactive}
Matthew Hausknecht, Prithviraj Ammanabrolu, Marc-Alexandre C{\^o}t{\'e}, and Xingdi Yuan. 2020.
\newblock Interactive fiction games: A colossal adventure.
\newblock In \emph{Proceedings of the AAAI Conference on Artificial Intelligence}, volume~34, pages 7903--7910.

\bibitem[{He et~al.(2024)He, Zhang, Yan, Wu, and Chen}]{he2024ideaenhancingrulelearning}
Kaiyu He, Mian Zhang, Shuo Yan, Peilin Wu, and Zhiyu~Zoey Chen. 2024.
\newblock \href {https://arxiv.org/abs/2408.10455} {Idea: Enhancing the rule learning ability of large language model agent through induction, deduction, and abduction}.
\newblock \emph{Preprint}, arXiv:2408.10455.

\bibitem[{Ke et~al.(2024)Ke, Cai, Jahangard, Wang, Haghighi, and Rezatofighi}]{ke2024hydra}
Fucai Ke, Zhixi Cai, Simindokht Jahangard, Weiqing Wang, Pari~Delir Haghighi, and Hamid Rezatofighi. 2024.
\newblock \href {https://doi.org/10.1007/978-3-031-72661-3_8} {Hydra: A hyper agent for dynamic compositional visual reasoning}.
\newblock In \emph{European Conference on Computer Vision}, pages 132--149. Springer.

\bibitem[{Kim et~al.(2024)Kim, Min, Kim, Kim, Jeung, and Choi}]{kim2024realfred}
Taewoong Kim, Cheolhong Min, Byeonghwi Kim, Jinyeon Kim, Wonje Jeung, and Jonghyun Choi. 2024.
\newblock Realfred: An embodied instruction following benchmark in photo-realistic environments.
\newblock In \emph{European Conference on Computer Vision}, pages 346--364. Springer.

\bibitem[{Kolve et~al.(2017)Kolve, Mottaghi, Han, VanderBilt, Weihs, Herrasti, Gordon, Zhu, Gupta, and Farhadi}]{ai2thor}
Eric Kolve, Roozbeh Mottaghi, Winson Han, Eli VanderBilt, Luca Weihs, Alvaro Herrasti, Daniel Gordon, Yuke Zhu, Abhinav Gupta, and Ali Farhadi. 2017.
\newblock {AI2-THOR: An Interactive 3D Environment for Visual AI}.
\newblock \emph{arXiv}.

\bibitem[{Li et~al.(2024{\natexlab{a}})Li, Zhang, Guo, Zhang, Li, Zhang, Zhang, Zhang, Li, Liu et~al.}]{li2024llava}
Bo~Li, Yuanhan Zhang, Dong Guo, Renrui Zhang, Feng Li, Hao Zhang, Kaichen Zhang, Peiyuan Zhang, Yanwei Li, Ziwei Liu, and 1 others. 2024{\natexlab{a}}.
\newblock Llava-onevision: Easy visual task transfer.
\newblock \emph{arXiv preprint arXiv:2408.03326}.

\bibitem[{Li et~al.(2021)Li, Xia, Mart{\'\i}n-Mart{\'\i}n, Lingelbach, Srivastava, Shen, Vainio, Gokmen, Dharan, Jain et~al.}]{li2021igibson}
Chengshu Li, Fei Xia, Roberto Mart{\'\i}n-Mart{\'\i}n, Michael Lingelbach, Sanjana Srivastava, Bokui Shen, Kent Vainio, Cem Gokmen, Gokul Dharan, Tanish Jain, and 1 others. 2021.
\newblock igibson 2.0: Object-centric simulation for robot learning of everyday household tasks.
\newblock \emph{arXiv preprint arXiv:2108.03272}.

\bibitem[{Li et~al.(2024{\natexlab{b}})Li, Liu, Zhang, Wang, Xue, and Cai}]{li2024enhancing}
Zhiyuan Li, Dongnan Liu, Chaoyi Zhang, Heng Wang, Tengfei Xue, and Weidong Cai. 2024{\natexlab{b}}.
\newblock \href {https://aclanthology.org/2024.emnlp-main.114/} {Enhancing advanced visual reasoning ability of large language models}.
\newblock In \emph{Proceedings of the 2024 Conference on Empirical Methods in Natural Language Processing}, pages 1915--1929, Miami, Florida, USA. Association for Computational Linguistics.

\bibitem[{Liu et~al.(2023{\natexlab{a}})Liu, Li, Wu, and Lee}]{liu2023visual}
Haotian Liu, Chunyuan Li, Qingyang Wu, and Yong~Jae Lee. 2023{\natexlab{a}}.
\newblock Visual instruction tuning.
\newblock \emph{Advances in neural information processing systems}, 36:34892--34916.

\bibitem[{Liu et~al.(2024)Liu, Li, Zhang, Cui, Fang, Zheng, Zheng, and Song}]{Odyssey2024}
Shunyu Liu, Yaoru Li, Kongcheng Zhang, Zhenyu Cui, Wenkai Fang, Yuxuan Zheng, Tongya Zheng, and Mingli Song. 2024.
\newblock Odyssey: Empowering agents with open-world skills.
\newblock \emph{arXiv preprint arXiv:2407.15325}.

\bibitem[{Liu et~al.(2023{\natexlab{b}})Liu, Li, Zhang, Huang, Zha, and Huang}]{liu2023matcr}
Yiting Liu, Liang Li, Beichen Zhang, Shan Huang, Zheng-Jun Zha, and Qingming Huang. 2023{\natexlab{b}}.
\newblock \href {https://doi.org/10.1145/3581783.3612268} {{MaTCR}: Modality-aligned thought chain reasoning for multimodal task-oriented dialogue generation}.
\newblock In \emph{Proceedings of the 31st ACM International Conference on Multimedia}, pages 5776--5785. Association for Computing Machinery.

\bibitem[{{Manolis Savva*} et~al.(2019){Manolis Savva*}, {Abhishek Kadian*}, {Oleksandr Maksymets*}, Zhao, Wijmans, Jain, Straub, Liu, Koltun, Malik, Parikh, and Batra}]{habitat19iccv}
{Manolis Savva*}, {Abhishek Kadian*}, {Oleksandr Maksymets*}, Yili Zhao, Erik Wijmans, Bhavana Jain, Julian Straub, Jia Liu, Vladlen Koltun, Jitendra Malik, Devi Parikh, and Dhruv Batra. 2019.
\newblock Habitat: {A} {P}latform for {E}mbodied {AI} {R}esearch.
\newblock In \emph{Proceedings of the IEEE/CVF International Conference on Computer Vision (ICCV)}.

\bibitem[{Mnih et~al.(2013)Mnih, Kavukcuoglu, Silver, Graves, Antonoglou, Wierstra, and Riedmiller}]{mnih2013playing}
Volodymyr Mnih, Koray Kavukcuoglu, David Silver, Alex Graves, Ioannis Antonoglou, Daan Wierstra, and Martin Riedmiller. 2013.
\newblock Playing atari with deep reinforcement learning.
\newblock \emph{arXiv preprint arXiv:1312.5602}.

\bibitem[{Nagabandi et~al.(2019)Nagabandi, Clavera, Liu, Fearing, Abbeel, Levine, and Finn}]{nagabandi2019learning}
Anusha Nagabandi, Ignasi Clavera, Simin Liu, Ronald~S. Fearing, Pieter Abbeel, Sergey Levine, and Chelsea Finn. 2019.
\newblock \href {https://arxiv.org/abs/1803.11347} {Learning to adapt in dynamic, real-world environments through meta-reinforcement learning}.
\newblock In \emph{International Conference on Learning Representations (ICLR)}.

\bibitem[{Paglieri et~al.(2024)Paglieri, Cupia{\l}, Coward, Piterbarg, Wo{\l}czyk, Khan, Pignatelli, Kuci{\'n}ski, Pinto, Fergus, Foerster, Parker-Holder, and Rockt{\"a}schel}]{paglieri2024balrog}
Davide Paglieri, Bart{\l}omiej Cupia{\l}, Sam Coward, Ulyana Piterbarg, Maciej Wo{\l}czyk, Akbir Khan, Eduardo Pignatelli, {\L}ukasz Kuci{\'n}ski, Lerrel Pinto, Rob Fergus, Jakob~Nicolaus Foerster, Jack Parker-Holder, and Tim Rockt{\"a}schel. 2024.
\newblock Benchmarking agentic llm and vlm reasoning on games.
\newblock \emph{arXiv preprint arXiv:2411.13543}.

\bibitem[{Pan et~al.(2023)Pan, Chan, Zou, Li, Basart, Woodside, Zhang, Emmons, and Hendrycks}]{pan2023rewards}
Alexander Pan, Jun~Shern Chan, Andy Zou, Nathaniel Li, Steven Basart, Thomas Woodside, Hanlin Zhang, Scott Emmons, and Dan Hendrycks. 2023.
\newblock Do the rewards justify the means? measuring trade-offs between rewards and ethical behavior in the machiavelli benchmark.
\newblock In \emph{International conference on machine learning}, pages 26837--26867. PMLR.

\bibitem[{Puig et~al.(2018)Puig, Ra, Boben, Li, Wang, Fidler, and Torralba}]{puig2018virtualhome}
Xavier Puig, Kevin Ra, Marko Boben, Jiaman Li, Tingwu Wang, Sanja Fidler, and Antonio Torralba. 2018.
\newblock Virtualhome: Simulating household activities via programs.
\newblock In \emph{Proceedings of the IEEE Conference on Computer Vision and Pattern Recognition}, pages 8494--8502.

\bibitem[{Qian et~al.(2024)Qian, Han, Luo, He, Chen, Zhang, Du, Yao, Yang, Zhang et~al.}]{qian2024escapebench}
Cheng Qian, Peixuan Han, Qinyu Luo, Bingxiang He, Xiusi Chen, Yuji Zhang, Hongyi Du, Jiarui Yao, Xiaocheng Yang, Denghui Zhang, and 1 others. 2024.
\newblock Escapebench: Pushing language models to think outside the box.
\newblock \emph{arXiv preprint arXiv:2412.13549}.

\bibitem[{Shahbandeh et~al.(2024)Shahbandeh, Alian, Nashid, and Mesbah}]{shahbandeh2024naviqate}
Mobina Shahbandeh, Parsa Alian, Noor Nashid, and Ali Mesbah. 2024.
\newblock Naviqate: Functionality-guided web application navigation.
\newblock \emph{arXiv preprint arXiv:2409.10741}.

\bibitem[{Shi et~al.(2017)Shi, Karpathy, Fan, Hernandez, and Liang}]{shi2017world}
Tianlin Shi, Andrej Karpathy, Linxi Fan, Jonathan Hernandez, and Percy Liang. 2017.
\newblock World of bits: An open-domain platform for web-based agents.
\newblock In \emph{International Conference on Machine Learning}, pages 3135--3144. PMLR.

\bibitem[{Shin et~al.(2024)Shin, Kim, Kang, Zhang et~al.}]{shin2024socratic}
Suyeon Shin, Junghyun Kim, Gi-Cheon Kang, Byoung-Tak Zhang, and 1 others. 2024.
\newblock Socratic planner: Inquiry-based zero-shot planning for embodied instruction following.
\newblock \emph{arXiv preprint arXiv:2404.15190}.

\bibitem[{Shridhar et~al.(2020)Shridhar, Thomason, Gordon, Bisk, Han, Mottaghi, Zettlemoyer, and Fox}]{Shridhar_2020_CVPR}
Mohit Shridhar, Jesse Thomason, Daniel Gordon, Yonatan Bisk, Winson Han, Roozbeh Mottaghi, Luke Zettlemoyer, and Dieter Fox. 2020.
\newblock Alfred: A benchmark for interpreting grounded instructions for everyday tasks.
\newblock In \emph{Proceedings of the IEEE/CVF Conference on Computer Vision and Pattern Recognition (CVPR)}.

\bibitem[{Shridhar et~al.()Shridhar, Yuan, Cote, Bisk, Trischler, and Hausknecht}]{shridharalfworld}
Mohit Shridhar, Xingdi Yuan, Marc-Alexandre Cote, Yonatan Bisk, Adam Trischler, and Matthew Hausknecht.
\newblock Alfworld: Aligning text and embodied environments for interactive learning.
\newblock In \emph{International Conference on Learning Representations}.

\bibitem[{Team et~al.(2024)Team, Georgiev, Lei, Burnell, Bai, Gulati, Tanzer, Vincent, Pan, Wang et~al.}]{team2024gemini}
Gemini Team, Petko Georgiev, Ving~Ian Lei, Ryan Burnell, Libin Bai, Anmol Gulati, Garrett Tanzer, Damien Vincent, Zhufeng Pan, Shibo Wang, and 1 others. 2024.
\newblock Gemini 1.5: Unlocking multimodal understanding across millions of tokens of context.
\newblock \emph{arXiv preprint arXiv:2403.05530}.

\bibitem[{Wang et~al.()Wang, Xie, Jiang, Mandlekar, Xiao, Zhu, Fan, and Anandkumar}]{wangvoyager}
Guanzhi Wang, Yuqi Xie, Yunfan Jiang, Ajay Mandlekar, Chaowei Xiao, Yuke Zhu, Linxi Fan, and Anima Anandkumar.
\newblock Voyager: An open-ended embodied agent with large language models.
\newblock \emph{Transactions on Machine Learning Research}.

\bibitem[{Wang et~al.(2022)Wang, Jansen, C{\^o}t{\'e}, and Ammanabrolu}]{wang2022scienceworld}
Ruoyao Wang, Peter Jansen, Marc-Alexandre C{\^o}t{\'e}, and Prithviraj Ammanabrolu. 2022.
\newblock Scienceworld: Is your agent smarter than a 5th grader?
\newblock In \emph{Proceedings of the 2022 Conference on Empirical Methods in Natural Language Processing}, pages 11279--11298.

\bibitem[{Wang et~al.(2025)Wang, Dong, Luo, Ruan, Cheng, Chen, Li, and Liu}]{wang2025multimodallargelanguagemodels}
Ziyue Wang, Yurui Dong, Fuwen Luo, Minyuan Ruan, Zhili Cheng, Chi Chen, Peng Li, and Yang Liu. 2025.
\newblock \href {https://arxiv.org/abs/2503.10042} {How do multimodal large language models handle complex multimodal reasoning? placing them in an extensible escape game}.
\newblock \emph{Preprint}, arXiv:2503.10042.

\bibitem[{Xu et~al.(2021)Xu, Masling, Du, Campagna, Heck, Landay, and Lam}]{xu2021grounding}
Nancy Xu, Sam Masling, Michael Du, Giovanni Campagna, Larry Heck, James Landay, and Monica~S Lam. 2021.
\newblock Grounding open-domain instructions to automate web support tasks.
\newblock \emph{arXiv preprint arXiv:2103.16057}.

\bibitem[{Yang et~al.(2024{\natexlab{a}})Yang, Yang, Hui, Zheng, Yu, Zhou, Li, Li, Liu, Huang, Dong, Wei, Lin, Tang, Wang, Yang, Tu, Zhang, Ma, Yang, Xu, Zhou, Bai, He, Lin, Dang, Lu, Chen, Yang, Li, Xue, Ni, Zhang, Wang, Peng, Men, Gao, Lin, Wang, Bai, Tan, Zhu, Li, Liu, Ge, Deng, Zhou, Ren, Zhang, Wei, Ren, Liu, Fan, Yao, Zhang, Wan, Chu, Liu, Cui, Zhang, Guo, and Fan}]{yang2024qwen2}
An~Yang, Baosong Yang, Binyuan Hui, Bo~Zheng, Bowen Yu, Chang Zhou, Chengpeng Li, Chengyuan Li, Dayiheng Liu, Fei Huang, Guanting Dong, Haoran Wei, Huan Lin, Jialong Tang, Jialin Wang, Jian Yang, Jianhong Tu, Jianwei Zhang, Jianxin Ma, and 43 others. 2024{\natexlab{a}}.
\newblock \href {https://arxiv.org/abs/2407.10671} {Qwen2 technical report}.
\newblock \emph{Preprint}, arXiv:2407.10671.

\bibitem[{Yang et~al.(2024{\natexlab{b}})Yang, Yang, Zhang, Hui, Zheng, Yu, Li, Liu, Huang, Wei et~al.}]{yang2024qwen2_5}
An~Yang, Baosong Yang, Beichen Zhang, Binyuan Hui, Bo~Zheng, Bowen Yu, Chengyuan Li, Dayiheng Liu, Fei Huang, Haoran Wei, and 1 others. 2024{\natexlab{b}}.
\newblock Qwen2. 5 technical report.
\newblock \emph{arXiv preprint arXiv:2412.15115}.

\bibitem[{Yao et~al.(2022)Yao, Chen, Yang, and Narasimhan}]{yao2022webshop}
Shunyu Yao, Howard Chen, John Yang, and Karthik Narasimhan. 2022.
\newblock Webshop: Towards scalable real-world web interaction with grounded language agents.
\newblock \emph{Advances in Neural Information Processing Systems}, 35:20744--20757.

\bibitem[{Yao et~al.(2023)Yao, Zhao, Yu, Du, Shafran, Narasimhan, and Cao}]{yao2023react}
Shunyu Yao, Jeffrey Zhao, Dian Yu, Nan Du, Izhak Shafran, Karthik Narasimhan, and Yuan Cao. 2023.
\newblock React: Synergizing reasoning and acting in language models.
\newblock In \emph{International Conference on Learning Representations (ICLR)}.

\bibitem[{Young et~al.(2024)Young, Chen, Li, Huang, Zhang, Zhang, Wang, Li, Zhu, Chen et~al.}]{young2024yi}
Alex Young, Bei Chen, Chao Li, Chengen Huang, Ge~Zhang, Guanwei Zhang, Guoyin Wang, Heng Li, Jiangcheng Zhu, Jianqun Chen, and 1 others. 2024.
\newblock Yi: Open foundation models by 01. ai.
\newblock \emph{arXiv preprint arXiv:2403.04652}.

\bibitem[{Zhang et~al.(2018)Zhang, Lu, and Zhou}]{zhang2018adaptive}
Lijun Zhang, Shuang Lu, and Zhi-Hua Zhou. 2018.
\newblock \href {https://proceedings.neurips.cc/paper/2018/file/10a5ab2db37feedfdeaab192ead4ac0e-Paper.pdf} {Adaptive online learning in dynamic environments}.
\newblock In \emph{Advances in Neural Information Processing Systems (NeurIPS)}, pages 1323--1333.

\bibitem[{Zhao et~al.(2023)Zhao, Huang, Xu, Lin, Liu, and Huang}]{zhao2023expel}
Andrew Zhao, Daniel Huang, Quentin Xu, Matthieu Lin, Yong-Jin Liu, and Gao Huang. 2023.
\newblock \href {https://arxiv.org/abs/2308.10144} {{ExpeL}: {LLM} agents are experiential learners}.
\newblock In \emph{Proceedings of the AAAI Conference on Artificial Intelligence}.

\bibitem[{Zhao et~al.(2024)Zhao, Zong, Zhang, and Hospedales}]{zhao2024benchmarking}
Bingchen Zhao, Yongshuo Zong, Letian Zhang, and Timothy Hospedales. 2024.
\newblock Benchmarking multi-image understanding in vision and language models: Perception, knowledge, reasoning, and multi-hop reasoning.
\newblock \emph{arXiv preprint arXiv:2406.12742}.

\bibitem[{Zheng et~al.(2025)Zheng, Shi, Cai, Li, Zhang, Li, Yu, and Ma}]{zheng2025lifelong}
Junhao Zheng, Chengming Shi, Xidi Cai, Qiuke Li, Duzhen Zhang, Chenxing Li, Dong Yu, and Qianli Ma. 2025.
\newblock Lifelong learning of large language model based agents: A roadmap.
\newblock \emph{arXiv preprint arXiv:2501.07278}.

\bibitem[{Zhou et~al.()Zhou, Xu, Zhu, Zhou, Lo, Sridhar, Cheng, Ou, Bisk, Fried et~al.}]{zhouwebarena}
Shuyan Zhou, Frank~F Xu, Hao Zhu, Xuhui Zhou, Robert Lo, Abishek Sridhar, Xianyi Cheng, Tianyue Ou, Yonatan Bisk, Daniel Fried, and 1 others.
\newblock Webarena: A realistic web environment for building autonomous agents.
\newblock In \emph{The Twelfth International Conference on Learning Representations}.

\bibitem[{Zhou et~al.(2024)Zhou, Ou, Ding, Li, Wu, Wang, Chen, Wang, Xu, Zhang, Chen, and Jiang}]{zhou2024symbolic}
Wangchunshu Zhou, Yixin Ou, Shengwei Ding, Long Li, Jialong Wu, Tiannan Wang, Jiamin Chen, Shuai Wang, Xiaohua Xu, Ningyu Zhang, Huajun Chen, and Yuchen~Eleanor Jiang. 2024.
\newblock \href {https://arxiv.org/abs/2406.18532} {Symbolic learning enables self-evolving agents}.
\newblock \emph{arXiv preprint arXiv:2406.18532}.

\bibitem[{Zhu et~al.(2023)Zhu, Chen, Tian, Tao, Su, Yang, Huang, Li, Lu, Wang, Qiao, Zhang, and Dai}]{zhu2023ghost}
Xizhou Zhu, Yuntao Chen, Hao Tian, Chenxin Tao, Weijie Su, Chenyu Yang, Gao Huang, Bin Li, Lewei Lu, Xiaogang Wang, Yu~Qiao, Zhaoxiang Zhang, and Jifeng Dai. 2023.
\newblock \href {https://arxiv.org/abs/2305.17144} {Ghost in the minecraft: Generally capable agents for open-world environments via large language models with text-based knowledge and memory}.
\newblock \emph{arXiv preprint arXiv:2305.17144}.

\end{thebibliography}

\newpage
\appendix

\section*{Appendix Overview}
\label{sec:appendix}

\begin{itemize}

\item In Appendix~\ref{sec:expdetails}, we provide details about our experimental setup and the prompt template used.

\item In Appendix~\ref{sec:datasetdetail}, we provide additional information about our data construction process.

\item In Appendix~\ref{sec:additional_experiment}, we provide results and analysis for additional experiments that were not covered in the main paper.

\item In Appendix~~\ref{sec:moredetail}, we provide additional details not extensively covered in the main paper.

\item In Appendix~\ref{sec:escape}, we briefly introduce and analyze works that address the topic of escape rooms.

\item In Appendix~\ref{sec:human_study}, we present details about the human study conducted.
\end{itemize}

\section{Experimental details}
\label{sec:expdetails}
\subsection{Model List}
In the list below, we list the versions of the LLMs used in our experiments. For GPT, Gemini, and Claude models, we use the officially released APIs, and for the others, we use the model versions available on Huggingface. 

\begin{itemize}
    \item Claude-3.5-Sonnet ~\cite{claude3}:\\
    \texttt{claude-3-5-sonnet-20241022}
    \item GPT-4o ~\cite{achiam2023gpt}:\\
    \texttt{gpt-4o-2024-08-06}
    \item GPT-4o-mini ~\cite{achiam2023gpt}:\\
    \texttt{gpt-4o-mini}
    \item gemini-1.5-Pro ~\cite{team2024gemini}:\\
    \texttt{Gemini-1.5-Pro}
    \item gemini-2.0-Flash ~\cite{google2024gemini}:\\
    \texttt{Gemini-2.0-Flash}
    \item Qwen2.5-VL-32B-Instruct~\cite{bai2025qwen2_5vl}:\\
    \texttt{Qwen/Qwen2.5-VL-32B-Instruct}

    \item LLaVA-v1.6-34B~\cite{liu2023visual}:\\
    \texttt{llava-hf/llava-v1.6-34b-hf}
    \item LLaVA-OneVision-Qwen2-7B~\cite{li2024llava}:\\
    \texttt{llava-hf/llava-onevision-qwen2-7b-ov-hf}

    \item InternVL2-40B~\cite{chen2024expanding}:\\
    \texttt{OpenGVLab/InternVL2-40B}
    \item InternVL2.5-38B~\cite{chen2024expanding}:\\
    \texttt{OpenGVLab/InternVL2\_5-38B}

    \item Qwen2-7B-Instruct~\cite{yang2024qwen2}:\\
    \texttt{Qwen/Qwen2-7B-Instruct}
    \item Qwen2.5-32B-Instruct~\cite{yang2024qwen2_5}:\\
    \texttt{Qwen/Qwen2.5-32B-Instruct}

    \item Yi-34B~\cite{young2024yi}:\\
    \texttt{NousResearch/Nous-Hermes-2-Yi-34B}

    \item DeepSeek-R1-Distill-Qwen-32B~\cite{guo2025deepseek}: \\
    \texttt{deepseek-ai/DeepSeek-R1-Distill-\\Qwen-32B}
\end{itemize}

\subsection{MLLMs and Corresponding Backbone LLMs}
As shown in Table~\ref{tab:socratic}, we used only the LLM that served as backbone language model of corresponding MLLM for controlled experiments. Base LLMs corresponding to each MLLMs used in Table~\ref{tab:socratic} are shown in Table~\ref{tab:corresponding_llm}.~\footnote{The documents specifying the LLMs selected as language backbone models are as follows:
\begin{itemize}
    \item InternVL2.5-38B: \\ \href{https://huggingface.co/OpenGVLab/InternVL2_5-38B}{https://huggingface.co/OpenGVLab/InternVL2\_5-38B}
    \item InternVL2-40B: \\ \href{https://huggingface.co/OpenGVLab/InternVL2-40B}{https://huggingface.co/OpenGVLab/InternVL2-40B}
    \item LLaVA-v1.6-34B: \\ \href{https://huggingface.co/llava-hf/llava-v1.6-34b-hf}{https://huggingface.co/llava-hf/llava-v1.6-34b-hf}
    \item LLaVA-OneVision-7B: \\ \href{https://huggingface.co/llava-hf/llava-onevision-qwen2-7b-ov-hf}{https://huggingface.co/llava-hf/llava-onevision-qwen2-7b-ov-hf}
\end{itemize}}
\begin{table}[!htp]\centering
\resizebox{0.7\linewidth}{!}{
\begin{tabular}{cc}\toprule
\textbf{Model Type} & \textbf{Model Name} \\ \midrule
VLM & InternVL2.5-38B \\
Base LLM & Qwen2.5-32B-Instruct \\
Base LLM (R1) & DeepSeek-R1-Distill-Qwen-32B \\
\midrule
VLM & InternVL2-40B \\
Base LLM & Yi-34B \\
\midrule
VLM & LLaVA-v1.6-34B \\
Base LLM & Yi-34B \\
\midrule
VLM & LLaVA-OneVision-7B \\
Base LLM & Qwen2-7B-Instruct \\
\bottomrule
\end{tabular}
}
\caption{MLLMs and their corresponding base LLMs.}\label{tab:corresponding_llm}
\scriptsize
\end{table}

\subsection{Computational Resources and Environments}
For closed-source models, we used the OpenAI API for running GPT-4o and GPT-4o-mini, and the Anthropic API for running Claude-3.5-Sonnet.

For open-source models, we downloaded models from Huggingface and served models using the vLLM server. We categorized the models as follows: small models (under 10B parameters), large models (30 to 40B parameters). For small models, we allocated a single NVIDIA RTX 4090 GPU per model. For large models, we allocated two or four NVIDIA A6000 GPUs per model.

All results reported in this paper are aggregated from three different runs conducted in 20 rooms. The time required to complete each room ranged from as little as 10 minutes (for models under 10B parameters, which often terminated early due to the maximum turn limit) to as long as 90 minutes (for Deepseek-R1-Distill-Qwen-32B, which required considerable time for reasoning). For closed-source models, GPT-4o and Claude-3.5-Sonnet cost approximately \textdollar 2–3 per room for three separate experiments, totaling about \textdollar 50 for a complete run.

\subsection{Hints}
\label{sec:hint}
To facilitate the evaluation of the agent's progression along its path, all game logic in \datasetname is designed to be sequential. Therefore, the next checkpoint the agent must achieve after its current one is predetermined, as is the hint to guide it there. Hint messages are provided only during Hint-guided experiments if the agent gets stuck for 30 or more steps (i.e., shows no change in checkpoint progression).

Here are two examples of hint messages.
\begin{itemize}
    \item \textbf{Object interaction}:  ``Use key to unlock the wardrobe at the south wall.''
    \item \textbf{Answer for lock}:  ``Solve the numberlock in the safe at the east wall. The answer is 8056.''
\end{itemize}

\subsection{Prompt templates}
Table~\ref{tab:prompt_trait},~\ref{tab:prompt_trait2},~\ref{tab:prompt_trait3},~\ref{tab:prompt_trait4},~\ref{tab:prompt_trait5} shows experiment prompts.

\section{Dataset Construction Process}
\label{sec:datasetdetail}

\subsection{Asset Creation Using Trimble SketchUp}
\begin{figure*}[h]
    \centering
    \includegraphics[width=\textwidth]{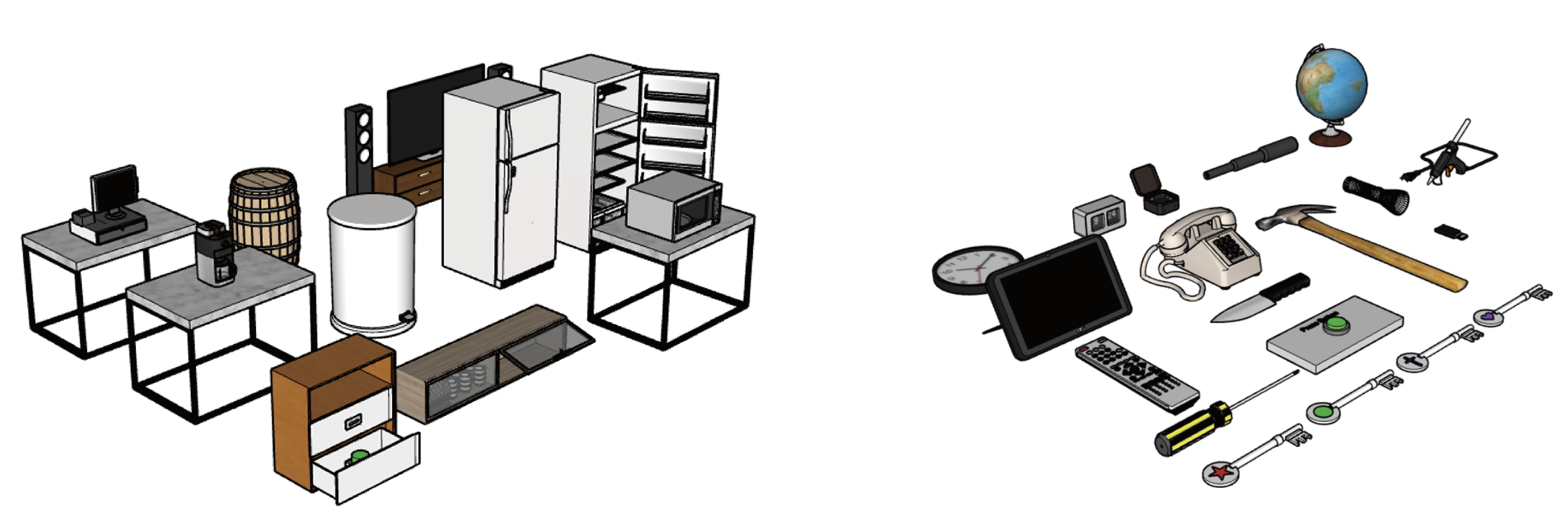}
    \caption{Objects created using Trimble Sketchup for VisEscape. On the left are objects of the \textit{Receptacle} type, and on the right are objects of the \textit{Item} type.}
    \label{fig:receptacle_sketchup}
\end{figure*}

We use Trimble SketchUp to create object assets (receptacles and items) for the room escape dataset. Receptacles are designed with predefined states, such as open/closed, locked/unlocked, and pushed/pulled, enabling agent interactions. To accommodate scene dynamics, we model these predefined states and generate corresponding receptacles to reflect those states. The visualized examples of receptacles and items are shown in Figure~\ref{fig:receptacle_sketchup}.

\begin{figure*}[h]
    \centering
    \includegraphics[width=\textwidth]{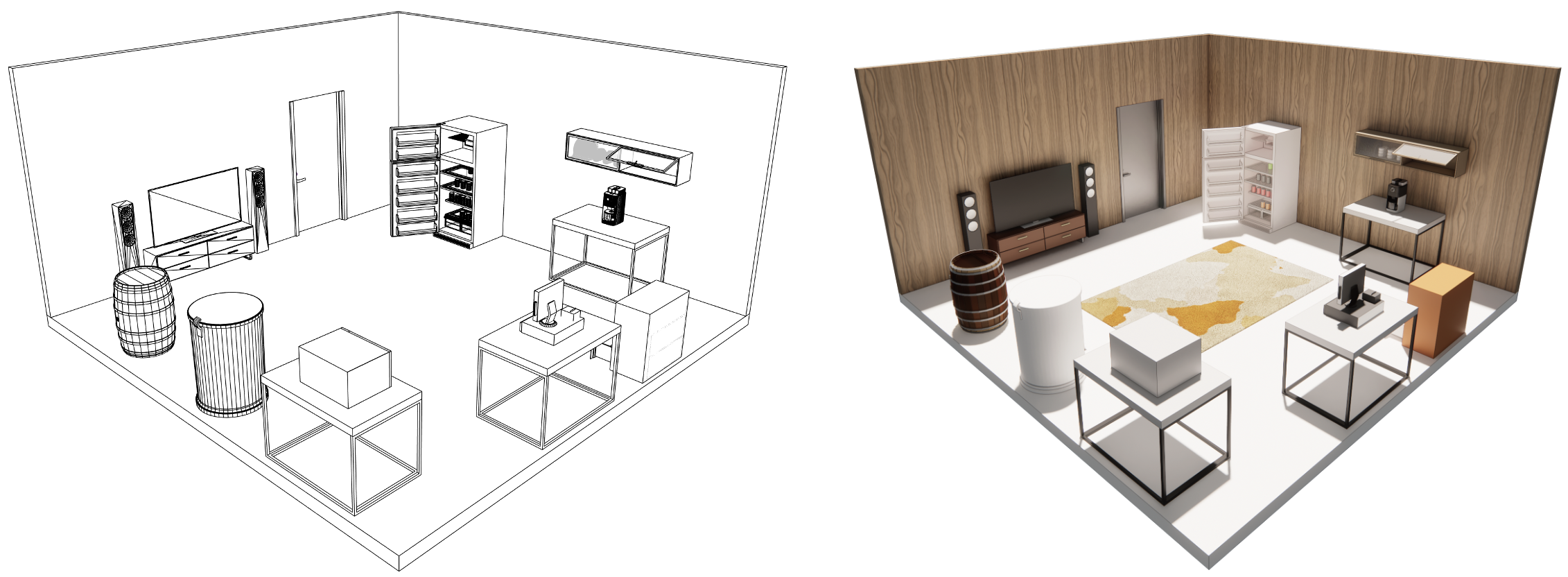}
    \caption{Comparison of applying visual rendering using Chaos Enscape. The left shows the view in Autodesk Revit before using Chaos Enscape, and the right shows the view in Chaos Enscape.}
    \label{fig:rendering_compare}
\end{figure*}

\subsection{Scene Creation Using Autodesk Revit}
We employ Autodesk Revit to construct room escape scenes by integrating the assets created in Trimble SketchUp. Autodesk Revit enables the precise placement of receptacles and objects within the scene to simulate realistic escape room environments. Each scene includes multiple viewpoints, which are categorized into three types:

\begin{itemize}
    \item \textbf{Wall View}: A view showing an entire wall on one side of east, west, north, or south. This view updates dynamically according to the states of receptacles and items belonging to each wall. 
    \item \textbf{Receptacle View}: A close-up view of a receptacle accessed through the ``inspect'' action from the Wall View. It changes to reflect the states of receptacles.
    \item \textbf{Item View}: A close-up view of an item accessed through the ``inspect'' action from the Receptacle View. It provides a close-up perspective of critical elements such as locks, puzzles, and other objects directly involved in problem solving.
\end{itemize}

\subsection{Enhancing Photo-realism Using Chaos Enscape}
\begin{figure*}[h]
    \centering
    \includegraphics[width=0.8\textwidth]{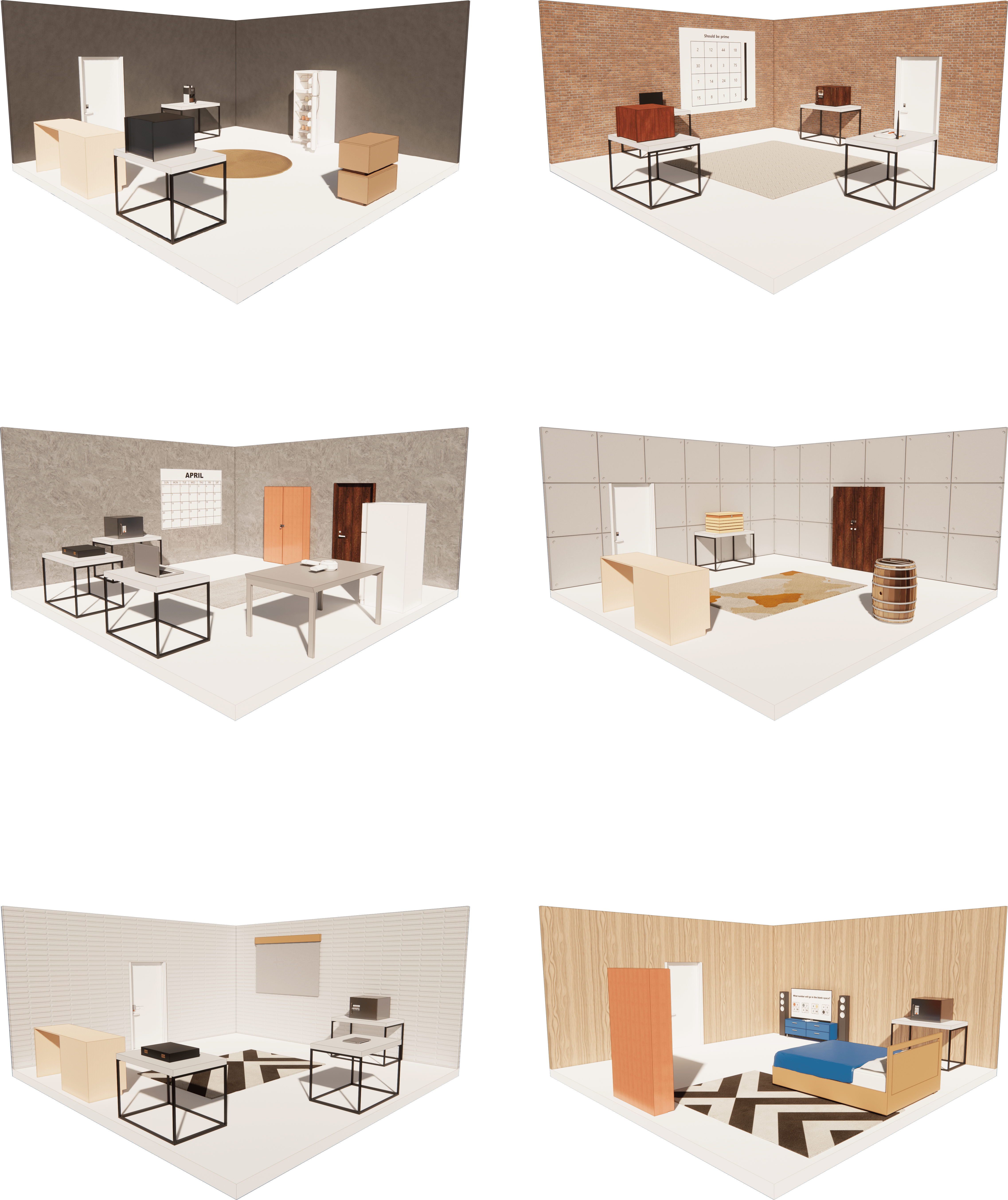}
    \caption{Example of rooms rendered using Chaos Enscape.}
    \label{fig:room_example}
\end{figure*}

To enhance the realism of the dataset, we use Chaos Enscape, a rendering tool that processes scene data generated in Autodesk Revit. Chaos Enscape improves the visual fidelity of the scene, providing photorealistic object representations. This increased realism facilitates effective agent reasoning and interaction within the environment, enabling accurate simulation and analysis of escape room scenarios. Refer to Figure~\ref{fig:rendering_compare} to check the before and after application of Chaos Enscape. Visualization of other rooms is in Figure~\ref{fig:room_example}.

\subsection{Game Logic Design}
We used OpenAI's web interface GPT models to design efficient and automated game logic. The authors then manually reviewed the generated game logic design to ensure error-free gameplay. The prompt used for game logic design is in Table~\ref{tab:prompt_logicdesign}.

\section{Additional Experiments and Analysis}
\label{sec:additional_experiment}

\subsection{Tendency of Brute-Force Approaches}
\label{sec:brute-force}
We found that when attempting to solve visual quizzes, smaller models frequently resorted to brute-force strategies—randomly trying different passwords or sequences without grounding their actions in any meaningful clues. For example, for problems requiring agents to arrange a given set of numbers in the correct order, some agents repeatedly attempted various possible combinations. To verify whether this behavior constituted brute-force attempts, we directly annotated whether the reasoning process and the corresponding attempted-answer pairs indicated a brute-force approach or not. Specifically, we annotated actions according to the following rules: (1) If the reasoning was grounded in meaningful clues or observations, we did not annotate it as brute-force. (2) If the attempted answers involved repeatedly trying different arrangements or combinations without valid justification, then we annotated them as brute-force.

Table~\ref{tab:brute_force_analysis} demonstrates that while Claude-3.5-Sonnet and GPT-4o typically grounded their puzzle-solving attempts in meaningful visual clues, 
GPT-4o-mini, InternVL2.5-38B, and LLaVA-v1.6-34B frequently resorted to brute-force approaches. Additionally, models that made a greater effort to identify reasoning justifications for their answers exhibited higher success rates in actual experiments. This behavior partially explains why these lower-performing models consistently achieved lower SPL and step efficiency scores—they made more attempts and failed more frequently on quizzes.
\begin{table}[!htp]
\centering
\resizebox{\linewidth}{!}{
\begin{tabular}{cccccc}
\toprule
\textbf{Model}      & \textbf{Brute-force attempt ratio(\%)} & \textbf{Solved quiz ratio(\%)}\\
\midrule
Claude-3.5-Sonnet    & 0.0 & 21.2\\
GPT-4o                & 0.9& 25.9\\
GPT-4o-mini           & 10.7& 6.3\\
InternVL2.5-38B      & 10.5& 6.6\\
LLaVA-v1.6-34B       & 70.0& 0.0\\
\midrule
\end{tabular}
}
\caption{Results for puzzle-solving. \textbf{Brute-force attempt ratio} denotes the ratio of brute-force attempts among all answers submitted for numeric or alphabetic lock puzzles. \textbf{Solved quiz ratio} denotes the ratio of correctly solved locks to the total number of locks in \datasetname.}
\label{tab:brute_force_analysis}
\end{table}

\subsection{Reasoning and Visual quizzes}
\label{sec:reason_quiz}
In \datasetname, all visual quizzes require associative thinking abilities. Analyzing game trajectories of diverse models, we find that most failures were on visual quizzes. So, we investigated whether these were problems they inherently couldn't solve, or if they failed because they couldn't properly associate the key clues within the escape room environment. For this,  we reformed them as multi-image Visual Question Answering (VQA) problems. By providing the model with both the clue image and the target lock image, we inform them that the two images are related to each other, and then have them solve the problem. Used prompts are in Table~\ref{tab:prompt_multiimage_vqa}.

Furthermore, we compare pass@1 and pass@10 to determine if models could correctly solve the task via iterative reasoning, even when their initial attempt failed. To support this repeated hypothesizing and testing, the model's previously generated reasoning sentence for an incorrect answer, along with the incorrect answer itself, was fed back as input for the subsequent attempt.

We selected one visual quiz from each room and then evaluated the performance of each model. Table~\ref{tab:reasoning_quiz} shows the number of correct answers by each model on these 20 visual quizzes. Results show that open-source models generally demonstrate lower performance (Pass@1 without reasoning) on these visual quizzes compared to frontier models. Also, Reasoning tends to improve the number of correctly solved quizzes across all models. 

While frontier models like GPT-4o and Claude-3.5-Sonnet show substantial performance gains with reasoning, the extent of this improvement varies, with some open-source models also showing notable gains, particularly when more attempts are allowed (e.g., InternVL2.5-38B shows a +5 improvement in Pass@10 with reasoning).

Also, iterative reasoning, as indicated by the general increase in scores from Pass@1 to Pass@10 under reasoning conditions, leads to a higher number of correct answers. This suggests that an iterative process of hypothesis formulation and testing enables the models to reach the correct solutions.

\begin{table}[!htp]\centering

\scriptsize
\begin{tabular}{lcccc}\toprule
\multirow{2}{*}{Model} &\multicolumn{2}{c}{\textbf{Pass@1}} &\multicolumn{2}{c}{\textbf{Pass@10}} \\\cmidrule{2-5}
&No Reason &Reason &No Reason &Reason \\\midrule
\gpt &9 &13 (+4) &9 &14 (+5) \\
\claude &6 &10 (+4) &11 &14 (+3) \\
\geminipro &7 &8 (+1) &9 &10 (+1) \\
\gptmini &6 &7 (+1) &7 &8 (+1) \\
\qwenvl &3 &5 (+2) &4 &6 (+2) \\
\internfive &2 &2 (+0) &3 &8 (+5) \\
\interntwo &2 &2 (+0) &2 &5 (+3) \\
\bottomrule
\end{tabular}
\caption{Comparison of pass@1 and pass@10 scores for models: performance with and without reasoning.}\label{tab:reasoning_quiz}
\end{table}

\subsection{Effect of the Exploration Memory}

\begin{figure}[htbp]
\centering
\includegraphics[width=\linewidth]{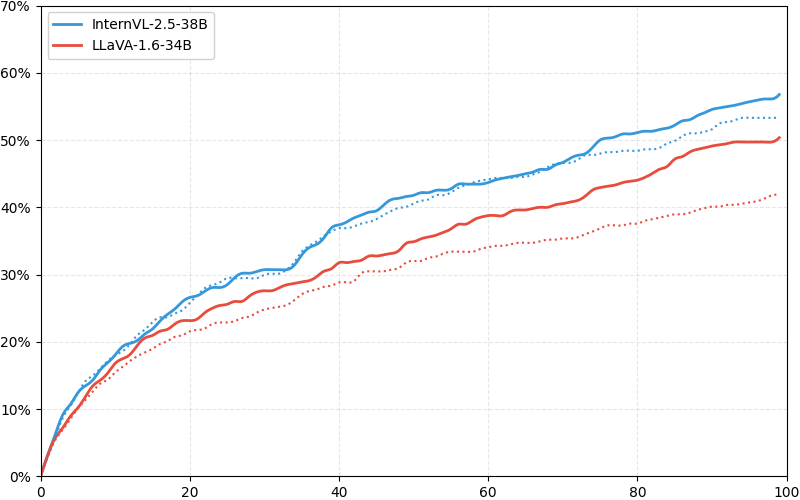}
\caption{Proportion of essential scenes observed by the agent at least once within the initial 100 steps. Solid and dotted lines denote results with and without exploration memory, respectively. Removing exploration memory significantly reduces the number of observed essential scenes in both models.}
\label{fig:exploration}
\end{figure}

We hypothesize that among the three components of memory management module, particularly the Exploration memory component, which tracks already explored versus unexplored scenes or objects, encourages agents to seek out unseen scenes or objects. To verify this, we conducted an additional ablation experiment in which we specifically removed the Exploration memory, while keeping the other two components (Structured spatial memory and Salient action memory) intact within the Memory module. For both settings, we measured the proportion of scenes observed by the agent from the set of essential observations (as determined by the oracle trajectory) required to successfully complete the game. Figure~\ref{fig:exploration} shows that InternVL2.5-38B exhibited only a slight difference in exploration of unique scenes, whereas LLaVA-v1.6-34B shows a substantial decrease.

\subsection{Comparison on Progress}

\begin{figure}[!htbp]
\centering
\includegraphics[width=\linewidth]{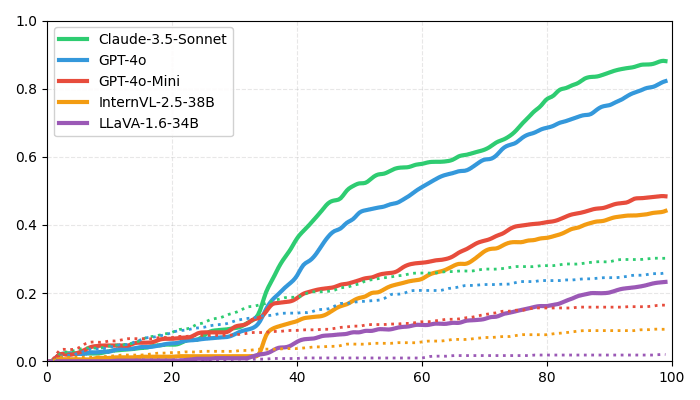}
\caption{Average GC (Goal Completion) progression across all rooms and experiments during the first 100 steps. The X-axis represents the number of steps, and the Y-axis represents the goal completion at each step. Solid lines represent results from hint-guided experiment, while dotted lines represent those without any hints.}
\label{fig:gcgraph}
\end{figure}

Figure~\ref{fig:gcgraph} illustrates the progression of goal completion across all rooms during the first 100 steps. Notably, while Claude-3.5-Sonnet performs best overall, it exhibits a distinctly accelerated trajectory after approximately 30-40 steps. This acceleration indicates that strong models better manage accumulated room-state changes, effectively leveraging previous observations to make strategic and efficient decisions later in the trajectory.

Additionally, models exhibiting rapid progress (Claude-3.5-Sonnet and GPT-4o) used fewer hints compared to other models (low HCR in Table~\ref{tab:main_hint}). This indicates that rapid progress by high-performing models does not stem from heavy reliance on hints; instead, they use hints selectively only to resolve specific obstacles and, once past these obstacles, accelerate progress by effectively leveraging information accumulated up to that point.

\subsection{Repetition}
\label{sec:repetition}
\begin{figure}[!htbp]
\centering
\includegraphics[width=\linewidth]{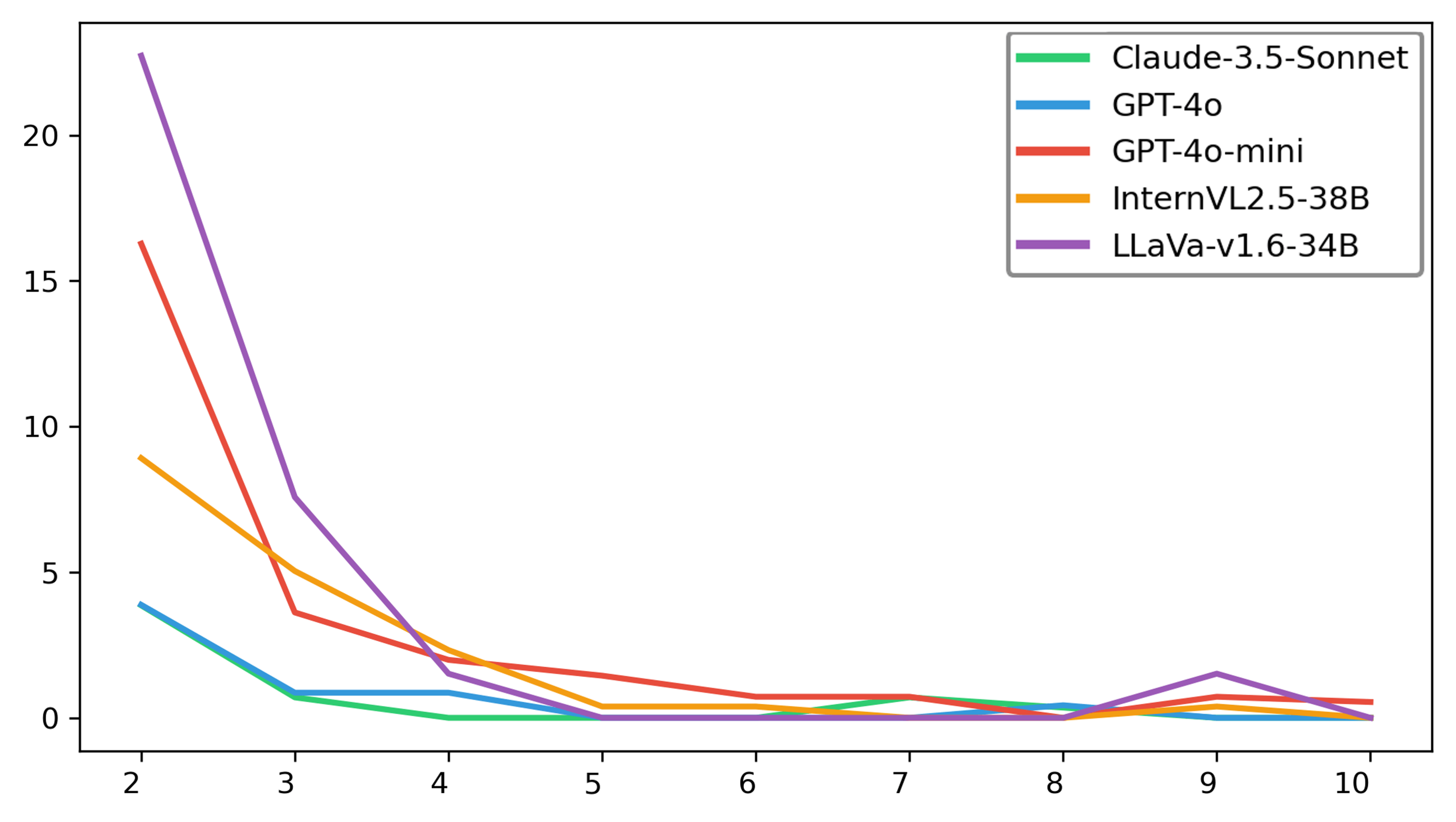}
\caption{The ratio of actions attempted multiple times (two or more repetitions) within a single trajectory, distributed across the frequency of attempt counts (ranging from 2 to 10+ repetitions). The X-axis represents the number of repeated attempts, and the Y-axis shows the ratio of these attempts out of all actions.}

\label{fig:duplication}
\end{figure}

To analyze the performance difference between frontier models and open-source models, we analyzed repetitive actions (repeating the same action in identical situations) exhibited by each model within their trajectories. We observe that lower-performing models frequently repeat actions they have already taken throughout their trajectories, getting stuck in repetitive cycles. Figure~\ref{fig:duplication} shows that these redundant attempts are more prevalent in lower-performing models.

\subsection{Statistics for Image Captions}
Since the Memory module uses image captions to construct and manage long-term memory, it is crucial to evaluate how accurately the MLLM captures essential information about the current state of the scene. To evaluate captions, we define the ground-truth (GT) state as the complete set of all receptacles and items present in an image obtained from the internal game engine. We then parse the generated captions and examine whether they contain the information consistent with the GT state. Table~\ref{tab:caption} shows that captions generated by InternVL2.5-38B provide the most accurate information in the most concise way, even when compared to GPT-4o.

\begin{table}[!htp]\centering
\resizebox{0.8\linewidth}{!}{%
\begin{tabular}{cccc}\toprule
Model &Accuracy(\%) &Avg. Length \\\midrule
GPT-4o &73.9 &47.03 \\
GPT-4o-mini &72.8 &64.46 \\
LLaVA-v1.6-34B &63.8 &91.19 \\
InternVL2.5-38B &75.1 &38.50 \\
\bottomrule
\end{tabular}
}
\caption{Evaluation on captions generated by each VLMs. Accuracy indicates whether the caption accurately describes the current game state. Avg. Length denotes the average length of all captions generated by each model for every unique scene.}
\label{tab:caption}
\scriptsize

\end{table}

\begin{table*}[!htp]
\centering
\resizebox{\textwidth}{!}{
    \scriptsize
    \begin{tabular}{cccccccccccc}
    \toprule
    & & \multicolumn{4}{c}{\textbf{\expbase}} & \multicolumn{5}{c}{\exphint} \\
    \cmidrule(lr){3-6}\cmidrule(lr){7-11}
    
    \textbf{Model(VLM / LLM)} & \textbf{Input Modality} & \textbf{SR(\%)} & \textbf{GC(\%)} & \textbf{SPL(\%)} & \textbf{Step} & \textbf{SR(\%)} & \textbf{GC(\%)} & \textbf{HCR(\%)} & \textbf{SPL(\%)} & \textbf{Step} \\
    \midrule

    GPT-4o & Image &13.3 &29.02 &3.91 &276.2 &95.0 &97.39 &20.74 &26.83 &112.9 \\
    GPT-4o & Caption &10.0 &23.54 &3.42 &280.7 &\textbf{98.3} &\textbf{98.00} &\textbf{17.39} &\textbf{33.60} &\textbf{91.8} \\
    \midrule
    GPT-4o-mini & Image &6.7 &21.31 &1.05 &291.2 &46.7 &67.40 &32.16 &8.93 &228.2 \\
    GPT-4o-mini & Caption &\textbf{11.7} &\textbf{24.21} &\textbf{3.25} &\textbf{280.3} &\textbf{53.3} &\textbf{73.94} &33.93 &\textbf{9.76} &\textbf{225.6} \\
    \midrule

    InternVL2.5-38B & Image &0.0 &6.20 &0.0 &- &66.7 &79.23 &41.69 &12.23 &210.5 \\
    InternVL2.5-38B + Base LLM & Caption &1.7 &9.57 &0.67 &295.9 &35.0 &53.76 &38.14 &7.08 &245.3 \\
    InternVL2.5-38B + Base LLM~(R1) & Caption &\textbf{10.3} &\textbf{19.58} &\textbf{7.81} &\textbf{261.6} &\textbf{80.0} &\textbf{87.06} &\textbf{25.53} &\textbf{19.37} &\textbf{166.8} \\
    
    \midrule
    InternVL2-40B & Image &0.0 &4.91 &0.0 & - &8.3 &48.18 &61.61 &0.96 &291.1 \\
    InternVL2-40B + Base LLM & Caption &0.0 &3.88 &0.0 &- &\textbf{32.0} &\textbf{68.00} &\textbf{48.22} &\textbf{4.39} &\textbf{250.6} \\
    \midrule
    LLaVA-v1.6-34B & Image &0.0 &0.54 &0.0 &- &15.0 &34.79 &64.97 &1.47 &289.8 \\
    LLaVA-v1.6-34B + Base LLM & Caption &0.0 &\textbf{4.14} &0.0 &- &\textbf{38.3} &\textbf{61.80} &\textbf{50.08} &\textbf{4.59} &\textbf{257.7} \\
    \midrule
    LLaVA-OneVision-7B & Image &0.0 &1.25 &0.0 &- &8.3 &41.02 &62.89 &0.80 &294.9 \\
    LLaVA-OneVision-7B + Base LLM & Caption &0.0 &\textbf{1.62} &0.0 &- &\textbf{25.0} &\textbf{58.93} &70.70 &\textbf{2.40} &\textbf{287.3} \\
    
    \bottomrule
    \end{tabular}
}
\caption{Full results are shown in Table 4. For GPT-4o and GPT-4o-mini, we used the same model but varied the input modality type, since they do not provide language models that receives text input only.}
\label{tab:socratic_full}
\end{table*}

\subsection{Full Results of LLM Experiment}
Table~\ref{tab:socratic_full} includes results for GPT-4o and GPT-4o-mini, which were omitted in Table~\ref{tab:socratic}. When GPT-4o-mini used captions instead of images as input, there was a slight improvement across all metrics, but it did not show a significant difference compared to other open-source models. Notably, GPT-4o's performance decreased in \expbase~but slightly improved in~\exphint~setting.

\begin{table}[!htp]\centering

\scriptsize
\resizebox{\linewidth}{!}{%
\begin{tabular}{lcccc}\toprule
\multirow{2}{*}{Model}&\multicolumn{2}{c}{\expbase} &\multicolumn{2}{c}{\exphint} \\\cmidrule{2-5}
&Step(\textit{Mean.}) &Step(\textit{Std.}) &Step(\textit{Mean.}) &Step(\textit{Std.}) \\\midrule
\claude &249.5 &47.11 &103.1 &29.38 \\
\gpt &276.2 &67.54 &112.9 &38.59 \\
\gptmini &291.2 &54.82 &228.2 &45.98 \\
\geminipro &258.1 &84.44 &163.3 &41.45 \\
\geminiflash &255.0 &48.12 & 179.2&41.97 \\
\qwenvl &295.8 &- &197 &32.62 \\
\internfive &300.0 &12.5 &210.5 &52.26 \\
\interntwo &300.0 &- &291.1 &59.85 \\
\llava &300.0 &- &289.8 &51.51 \\
\llavaov &300.0 &- &294.9 &43.41 \\
\bottomrule
\end{tabular}}
\caption{Mean and standard deviation of gameplay steps for each model. The standard deviation was calculated using only successful escape trajectories. For models that failed to escape entirely, the standard deviation is marked with "-".}
\label{tab:step_std}
\end{table}

\subsection{Reasoning Decompsition}
\begin{figure}[h]
    \centering
    \includegraphics[width=\linewidth]{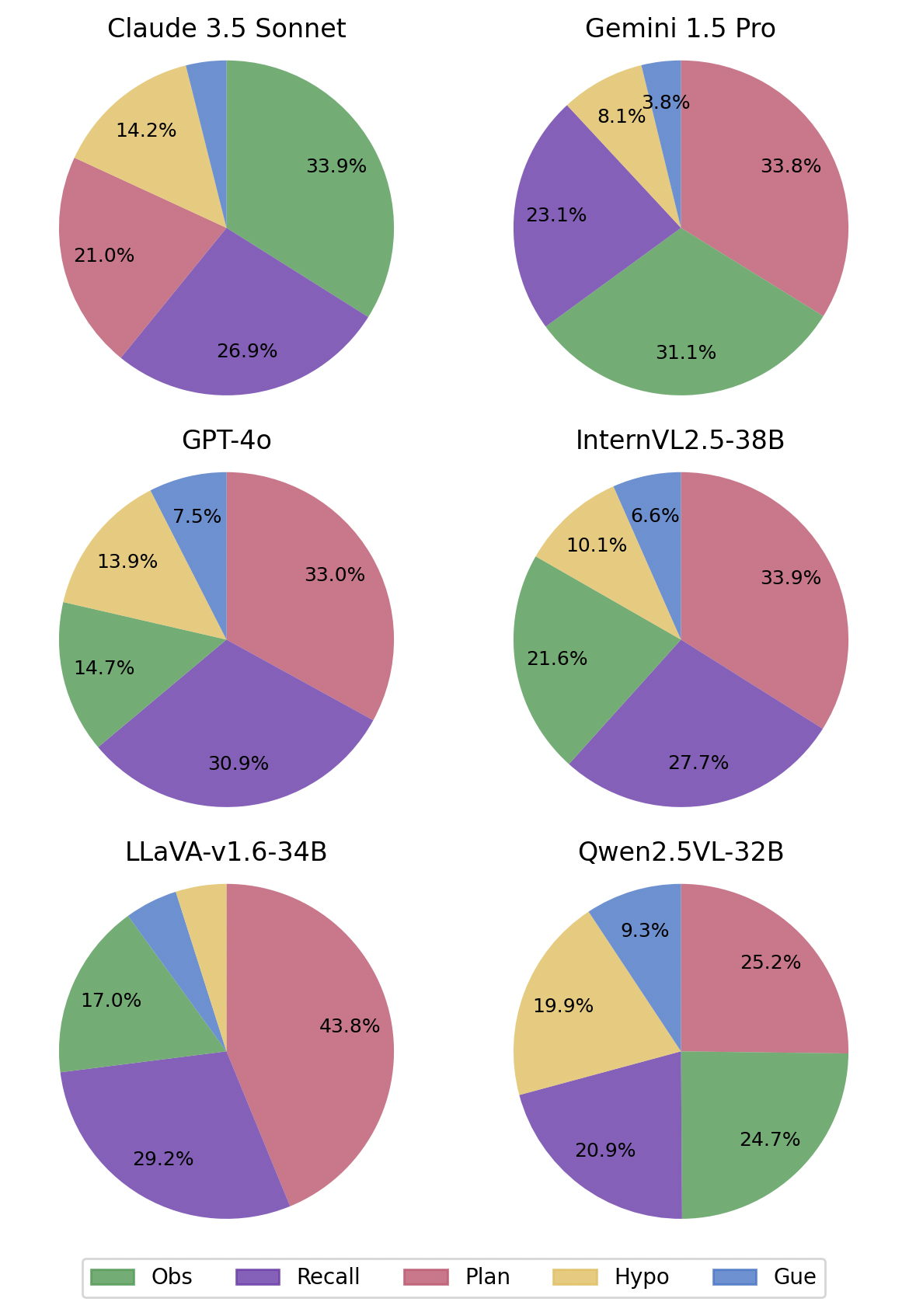}
    \caption{Proportion for each reasoning component for other models.}
    \label{fig:reasoning_component_full}
\end{figure}

The proportions of reasoning components and their composition for other models are shown in Table~\ref{tab:appendix_category_example} and Figure~\ref{fig:reasoning_component_full}.

\section{Detailed Explanation}
\label{sec:moredetail}

\subsection{Reasoning Component}
\label{sec:reasoning_component}
We have classified the reasoning processes that models undertake while solving problems in \datasetname as follows: Observation, Guess, Hypothesis Formation, Planning, and Recall. The classification criteria are as follows:

\begin{itemize}
    \item \textbf{Observation:} Visually perceiving, describing, or analyzing the current scene.
    \item \textbf{Guess:} Making a guess or assumption based on priors or knowledge, without exact evidence acquired in the environment.
    \item \textbf{Hypothesis Formation:} Forming a hypothesis or idea to test in the environment, based on any evidence acquired in the environment.
    \item \textbf{Planning:} Establishing or formulating action plans within the environment.
    \item \textbf{Recall:} Retrieving information from history - remembering the state/location of objects, actions previously performed, and other relevant details.
\end{itemize}

\begin{figure}[h]
    \centering
    \includegraphics[width=\linewidth]{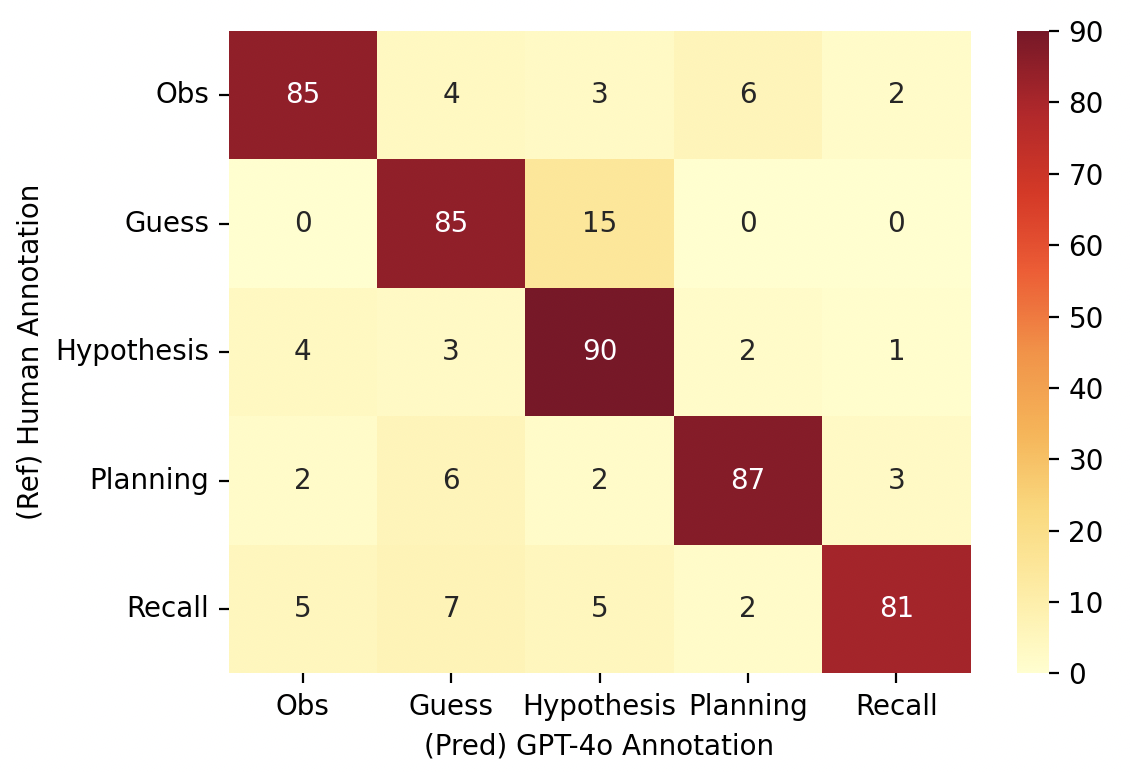}
    \caption{Heatmap of agreement between GPT-4o and human annotations for reasoning type classification.  In 85.6\% of the instances, the category annotations from GPT-4o and humans were identical.}
    \label{fig:category_annotation}
\end{figure}

We extract the pre-action reasoning and action performed by each model at every game step from the experimental logs, which were run with an applied memory management and a reasoning module. And they are annotated using GPT-4o~\cite{achiam2023gpt}. We set temperature as 0.0 for reproduction. To ensure the reliability of our machine-based evaluation method, we sampled 100 instances for each of the categories classified by GPT-4o, totaling 500 samples. These sentences were then manually annotated by humans following the same procedure. Figure~\ref{fig:category_annotation} shows a heatmap between human and GPT-4o annotations, suggesting that while the model's classifications largely align with human annotations, the model frequently misclassified the \textit{Guess} type as \textit{Hypothesis Formation}.

\begin{table}[!htp]\centering

\scriptsize
\resizebox{\linewidth}{!}{

\begin{tabular}{llccc}\toprule
\textbf{Model} &\textbf{Composition} &  \textbf{Ratio} & \textbf{Length}\\

\midrule
\multirow{5}{*}{\claude} & \textit{(Obs, Recall, Plan)} & \gradientcell{18.1} & 127.0 \\
 &\textit{(Obs, Recall, Hypo, Plan)}  &\gradientcell{15.8} & 141.5  \\
 &\textit{(Obs, Recall, Plan, Hypo)}  &\gradientcell{14.8} & 127.5  \\
 &\textit{(Obs, Recall, Obs, Hypo, Plan)}  &\gradientcell{9.5} & 155.2  \\
 &\textit{(Obs, Recall, Hypo)}  &\gradientcell{8.6} & 125.9  \\

\midrule
\multirow{5}{*}{\geminipro}  &\textit{(Obs, Plan)} &\gradientcell{28.8} & 28.1 \\
 &\textit{(Obs, Recall, Plan)}  &\gradientcell{18.6} & 46.1 \\
 &\textit{(Plan)}  &\gradientcell{17.4} & 6.5 \\
 &\textit{(Recall, Plan)}  &\gradientcell{17.0} & 32.8 \\
 &\textit{(Recall, Obs, Plan)}  &\gradientcell{6.0} & 43.0 \\

\midrule
\multirow{5}{*}{\gpt}  &\textit{(Recall, Plan)} &\gradientcell{31.6} & 62.8 \\
 &\textit{(Recall, Hypo)}  &\gradientcell{13.7} & 53.9 \\
 &\textit{(Recall, Hypo, Plan)}  &\gradientcell{11.0} & 67.4 \\
 &\textit{(Obs, Plan)}  &\gradientcell{8.8} & 54.0 \\
 &\textit{(Recall, Gue, Plan)}  &\gradientcell{8.5} & 65.5 \\

\midrule
\multirow{5}{*}{\internfive}  &\textit{(Recall, Plan)} &\gradientcell{23.3} & 76.7 \\
 &\textit{(Plan)}  &\gradientcell{20.9} & 13.1 \\
 &\textit{(Recall, Obs, Plan)}  &\gradientcell{11.5} & 90.7 \\
 &\textit{(Obs, Plan)}  &\gradientcell{9.6} & 68.7 \\
 &\textit{(Obs, Recall, Plan)}  &\gradientcell{8.4} & 93.6 \\

\midrule 
\multirow{5}{*}{\llava}  &\textit{(Recall, Plan)} &\gradientcell{38.5} & 55.3 \\
 &\textit{(Plan)}  &\gradientcell{21.3} & 39.4 \\
 &\textit{(Obs, Plan)}  &\gradientcell{11.7} & 54.4 \\
 &\textit{(Recall, Obs, Plan)}  &\gradientcell{8.9} & 64.0 \\
 &\textit{(Obs, Recall, Plan)}  &\gradientcell{6.5} & 67.0 \\

\midrule
\multirow{5}{*}{\qwenvl}  &\textit{(Obs, Recall, Hypo, Plan)} &\gradientcell{16.3} & 219.0 \\
 &\textit{(Obs, Recall, Plan)}  &\gradientcell{15.7} & 143.1 \\
 &\textit{(Recall, Obs, Hypo, Plan)}  &\gradientcell{15.1} & 143.1 \\
 &\textit{(Plan)}  &\gradientcell{12.7} & 6.4 \\
 &\textit{(Obs, Plan)}  &\gradientcell{7.9} & 95.0 \\
 
\bottomrule
 \end{tabular}}
\caption{Reasoning composition for other models.}
\label{tab:appendix_category_example}
\end{table}

\subsection{Memory Management Module}
\label{sec:memorymodule}

\begin{itemize}

\item \textbf{Structured spatial memory}: This memory hierarchically stores information about room walls, the receptacles on those walls, the states of these receptacles, along with their visual information and characteristics. This serves as a directional aid for navigating toward specific receptacles or items.

\item \textbf{Exploration memory}: This memory stores information about which objects and items have been inspected so far and which have not. This encourages the agent to explore unobserved information rather than getting stuck on the information acquired thus far.

\item \textbf{Salient action memory}: This memory selectively stores actions that are worth remembering, such as interactions with objects, attempting to enter passwords, etc. This allows the agent to make action decisions by referencing actions it has performed in the past.

\end{itemize}

\subsection{Visualization of Interactions in \datasetname}
\begin{figure}[!htbp]
\centering
\includegraphics[width=0.8\linewidth]{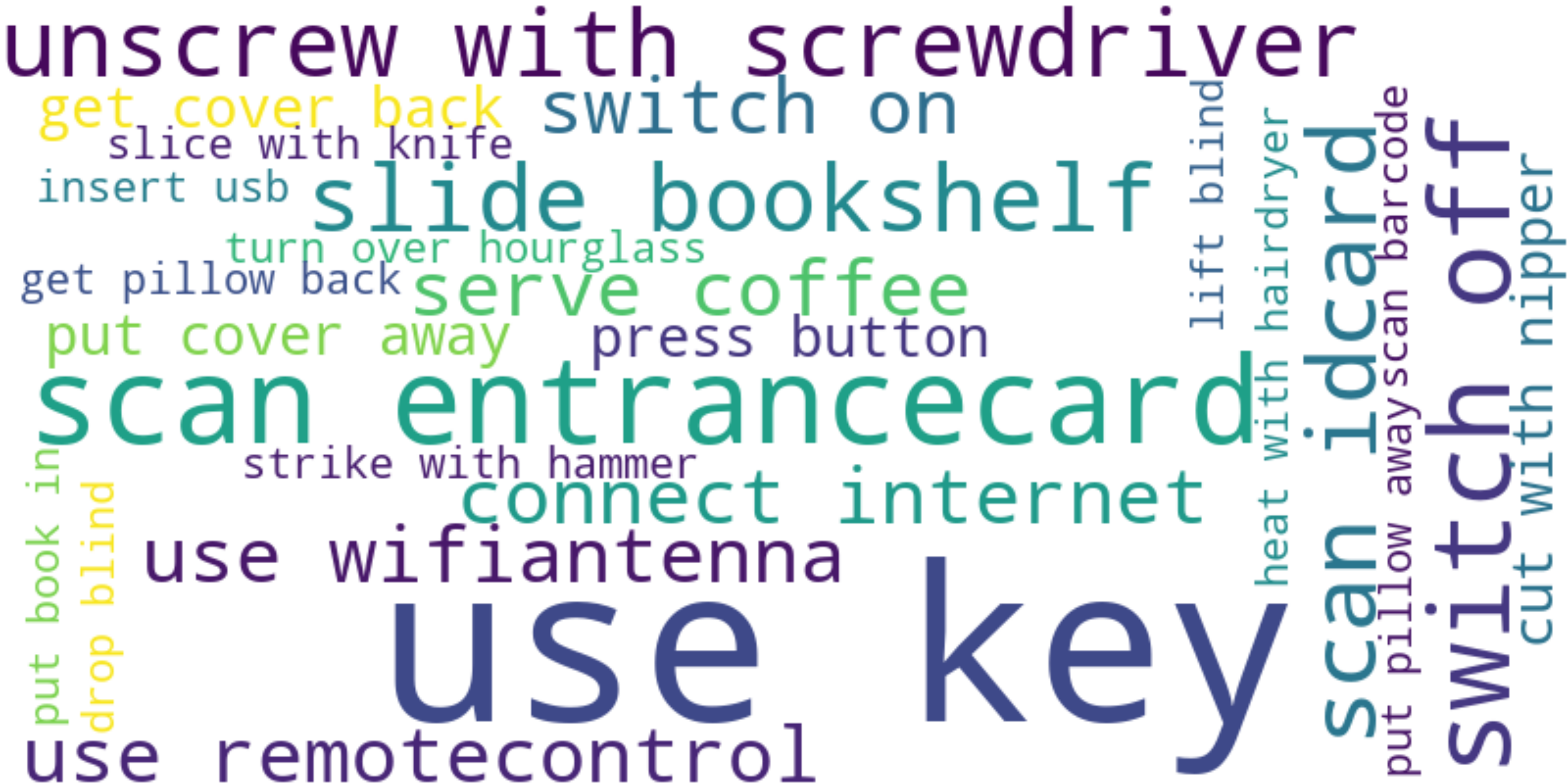}
\caption{Wordcloud of all types of actions defined in \datasetname. We excluded ``open'' and ``close'', which are available on most receptacles.}
\label{fig:wordcloud}
\end{figure}

Figure~\ref{fig:wordcloud} is a visualization of interactions defined in \datasetname. The size of each word is proportional to its frequency of appearance within the game. 

\subsection{Example of Visual Quizzes}
\begin{figure}[!htbp]
\centering
\includegraphics[width=\linewidth]{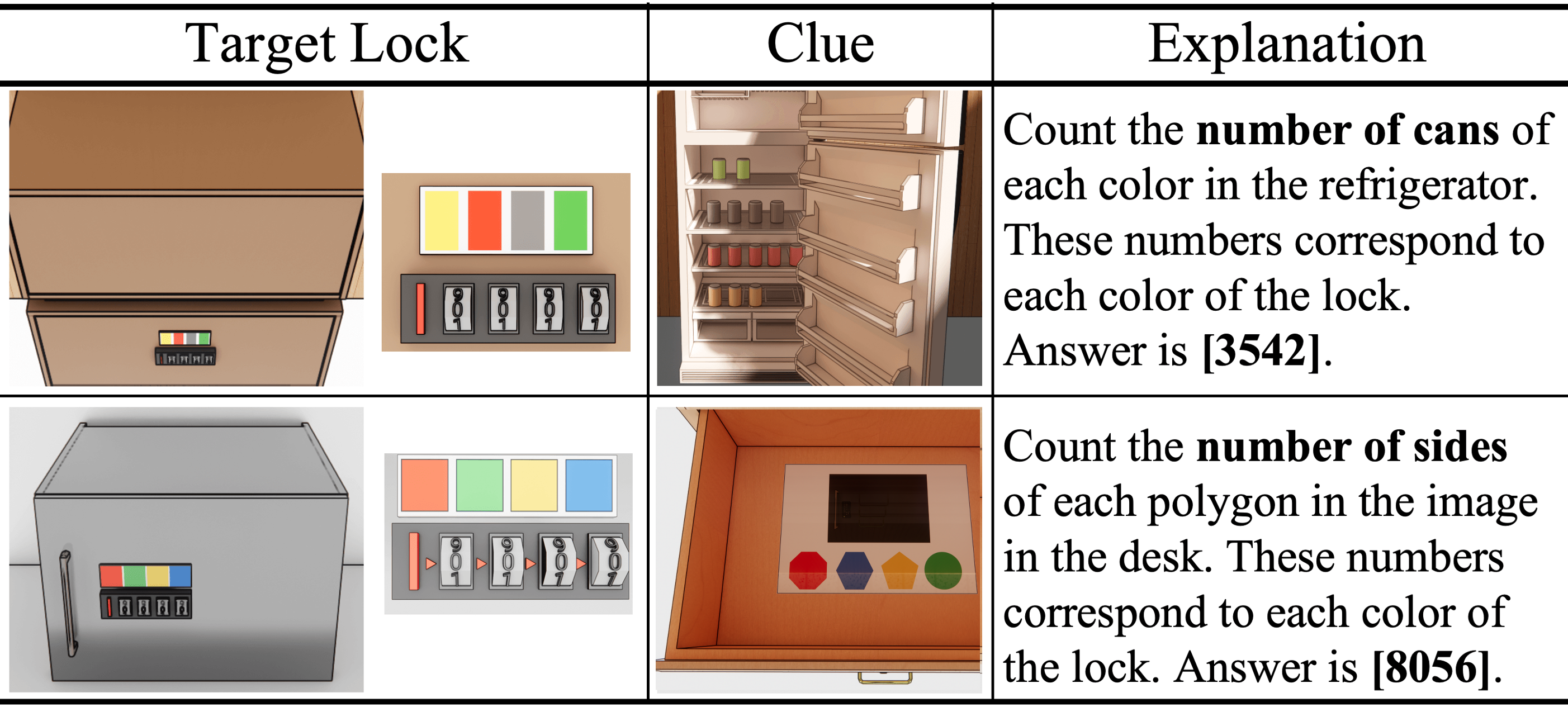}
\caption{Examples of visual quizzes that requires associative thinking in \datasetname. For all problems, we controlled the spatial positioning of the locks and clue images to ensure they remain distinctly separated. This design forces the agent not only to store each piece of information independently but also to identify relevant connections between them (such as ``color'' in the example above) and reason based on the established associations between these separate pieces of information.}
\label{fig:quiz}
\end{figure}

Figure~\ref{fig:quiz} shows two examples of visual quizzes in \datasetname.

\section{Related works on Escape Rooms}
\label{sec:escape}
EscapeBench~\cite{qian2024escapebench} proposes a benchmark to evaluate creative reasoning in text-only environments, with a focus on tool use as a proxy for measuring creativity. IDEA~\cite{he2024ideaenhancingrulelearning} presents an agent that mimics the loop of human rule-learning by integrating multiple forms of reasoning to guide its actions, focusing on solving text-based puzzles. MM-Escape~\cite{wang2025multimodallargelanguagemodels} aims to extend the escape room paradigm by building an embodied environment for the evaluation of MLLM agents. Departing from embodied setups, our visually grounded benchmark blends tool use with puzzle-solving to create a more flexible yet challenging reasoning framework for evaluating MLLMs.

\section{Human Study}
\label{sec:human_study}
\begin{table}[h]\centering
\resizebox{0.9\linewidth}{!}{
\begin{tabular}{c|c|c|c|c}
\toprule
Room & SR(\%) & GC(\%) &Steps  & Duration(s) \\
\midrule
1  & 75.0  &92.5& 41.8  & 231.3  \\
2  & 100.0  &100.0& 67.3  & 319.5  \\
3  & 50.0  &87.5& 71.1  & 319.5  \\
4  & 75.0  &93.8& 112.0 & 447.0  \\
5  & 100.0  &100.0& 77.3   & 274.5  \\
6  & 100.0  &100.0& 37.0    & 117.0  \\
7  & 100.0  &100.0& 46.8   & 225.0  \\
8  & 100.0  &100.0& 53.0  & 567.3  \\
9  & 75.0  &90.0& 52.0   & 567.3  \\
10  & 75.0  &91.7& 54.0    & 204.5  \\
11 & 50.0  &79.2& 37.8   & 231.3  \\
12 & 75.0  &92.9& 31.3  & 144.0  \\
13 & 100.0  &100.0& 41.5   & 199.8  \\
14 & 100.0  &100.0& 57.3    & 468.3  \\
15 & 100.0  &100.0& 36.0    & 151.0  \\
16 & 100.0  &100.0& 40.8  & 106.0  \\
17 & 25.0  &96.9& 63.3  & 488.3 \\
18 & 100.0  &100.0& 40.5 & 399.3  \\
19 & 75.0  &91.7& 43.8   & 378.5  \\
20 & 75.0  &100.0& 50.8  & 530.3  \\
\midrule
 Average& 82.5 & 95.8 & 52.8 & 318.5 \\
\bottomrule
\end{tabular}
}
\caption{Human evaluation per-room success rates, goal completions, step counts, and time durations.}
\label{tab:room_results}
\end{table}

\begin{figure*}[htbp]
\centering
\includegraphics[width=\textwidth]{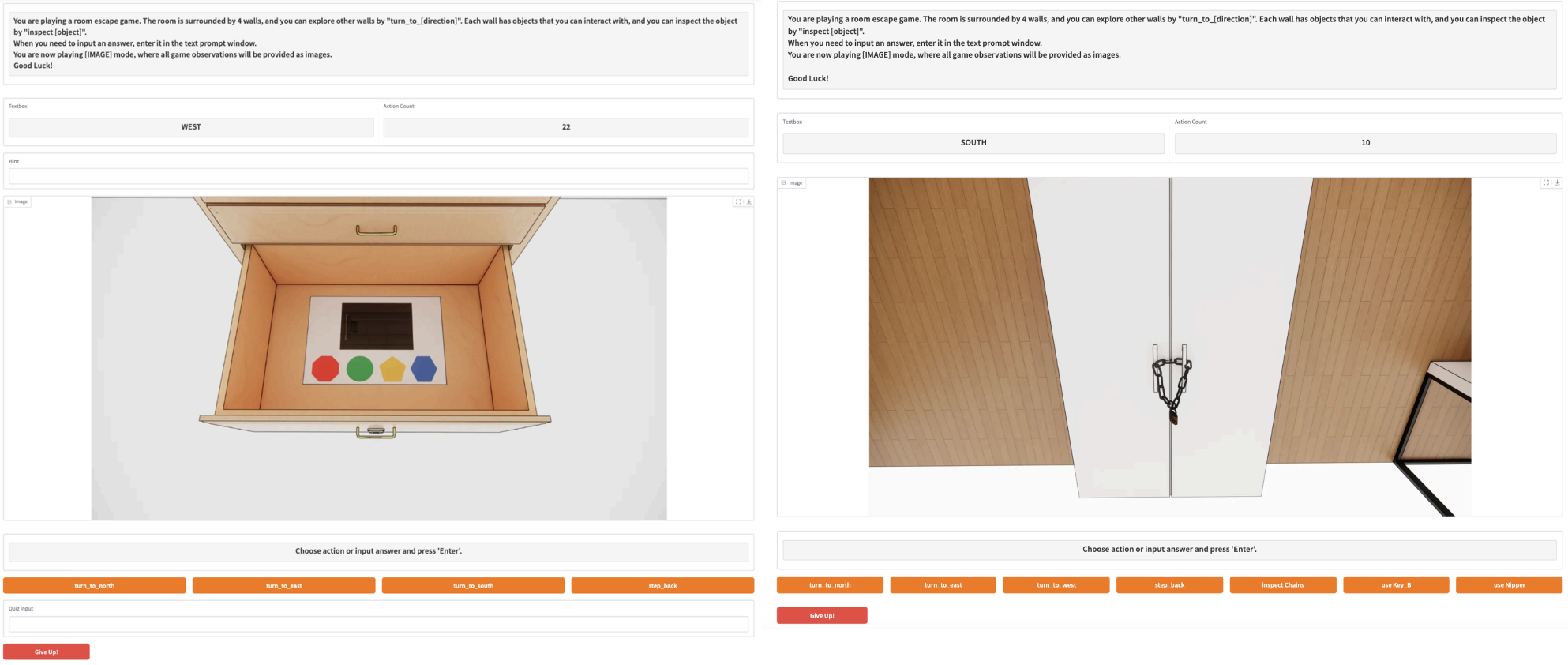}
\caption{User Interface generated by Gradio for conducting experiments for human trajectory.}
\label{fig:human_eval_gradio}
\end{figure*}

To compare the performance of AI models and human performance on escape room games, we conducted a human study. We recruited a total of 20 participants, including both individuals familiar and those unfamiliar with escape room games, and each was instructed to complete tasks in four different rooms. Figures~\ref{fig:human_eval_gradio} illustrate the Gradio interface used to provide the UI for playing \datasetname.

Table~\ref{tab:room_results} presents the results of the human evaluation for 20 rooms, including the success rate, goal completion, number of steps, and time duration for experiments. Participants achieved high success rates in most rooms. We paid them based on the minimum hourly wage.

\begin{table*}[!p]
\centering
\small

\begin{subfigure}{\textwidth}
\centering
\begin{tabular}{p{0.95\textwidth}}
\toprule
\textbf{Variables}: \\
\texttt{direction, available\_actions} \\

\midrule
\textbf{Initial prompt}: \\
\texttt{You are an AI agent playing a room escape game. The room is surrounded by 4 walls, and you can explore other walls by "turn\_to\_[direction]". Each wall has objects that you can interact with, and you can inspect the object by "inspect [object]". Please read the given information and task description below carefully and respond accordingly.} \\

\midrule
\textbf{Prompt}: \\
\{Initial Prompt\} \\
\texttt{<Current Observation>\{direction\} side of room - [IMAGE]</Current Observation>} \\
\texttt{Based on these information, choose next action to progress in the game. You can do one of the following actions: \{available\_actions\}} \\
\texttt{Before you act, think first, and then act. Your thought should be in section [THINK], and your action should be in section [ACTION]. In [ACTION], respond ONLY with the chosen action, no other text.} \\
\texttt{[THINK]} \\
\texttt{[ACTION]} \\

\bottomrule
\end{tabular}
\caption{Prompts for Reasoning module on initial step}
\label{tab:experiment-prompts-1}

\begin{tabular}{p{0.95\textwidth}}
\toprule
\textbf{Variables}: \\
\texttt{memory, salient\_action\_history, action\_history, direction, current\_scene\_desc, previous\_react, available\_actions, hint\_guideline\_text
} \\

\midrule
\textbf{Puzzle text} [OPTIONAL: if \textit{ispuzzle} mode]: \\
\texttt{<ANSWER> is an action to input the answer to open the lock you are facing. When you choose <ANSWER>, you should follow this format: \"<ANSWER>your answer</ANSWER>\".} \\

\midrule
\textbf{Prompt (Reason)}: \\
\{Initial Prompt\} \\
\texttt{<Memory>\{memory\}</Memory>} \\
\texttt{<Action Memory>\{salient\_action\_history\}</Action Memory>} \\
\texttt{<Recent actions(from oldest to latest)>\{action\_history\}</Recent actions>} \\
\texttt{<Current Observation>\{direction\} side of room - \{current\_scene\_desc\}</Current Observation>} \\
\texttt{<Your Thought and Action before this turn>\{previous\_react\}</Your Thought and Action before this turn>} \\
\texttt{\{available\_actions\}} \\
\{Puzzle text\} \\
\texttt{Before you act, think first, and then act. If there is a hint message, you should choose action to accomplish the guideline in hint message. \{hint\_guideline\_text\}} \\
\texttt{Your thought should be in section [THINK], and your action should be in section [ACTION]. In [ACTION], respond ONLY with the chosen action or <ANSWER>your answer</ANSWER>, no other text.} \\

\midrule
\textbf{Prompt (Retry)}: \\
\{Initial Prompt\} \\
\texttt{<Memory>\{memory\}</Memory>} \\
\texttt{<Your Previous Action> \{before\_action\}} \\
\texttt{<Available Actions> \{available\_actions\}} \\
\texttt{You just performed the <Your Previous Action>, but that action is not currently available in <Available Actions>. Referring to your memory, choose an action that is necessary to perform <Your Previous Action>.} \\
\texttt{\{hint\_guideline\_text\}} \\
\texttt{You should choose one of the following actions:\{available\_actions\}} \\
\texttt{Please respond following below format without any other text: [ACTION]} \\

\bottomrule
\end{tabular}
\caption{Prompts for Reasoning module and Retry}
\label{tab:experiment-prompts-2}

\end{subfigure}
\label{tab:templatetestend}

\caption{Prompts for the Reasoning module-before action.}
\label{tab:prompt_trait}
\end{table*}

\begin{table*}[!p]
\centering
\small

\begin{subfigure}{\textwidth}
\centering
\begin{tabular}{p{0.95\textwidth}}
\toprule
\textbf{Variables}: \\
\texttt{action\_history} \\

\midrule
\textbf{Initial prompt}: \\
\texttt{You are an AI agent playing a room escape game. The room is surrounded by 4 walls, and you can explore other walls by "turn\_to\_[direction]". Each wall has objects that you can interact with, and you can inspect the object by "inspect [object]". Please read the given information and task description below carefully and respond accordingly.} \\

\midrule
\textbf{last\_10\_history} (\texttt{for log in action\_history}) : \\
\texttt{Observation: [\{log['scene']\}]\textbackslash n} \\
\texttt{Action: [\{log['action']\}]-\{log['analysis']\}} \\

\midrule
\textbf{spatial\_json\_format}: \\
\texttt{\{"direction 1" : \{} \\
\texttt{\ \ \ "objects":["object1", "object2", ...]} \\
\texttt{\},} \\
\texttt{...\}} \\

\midrule
\textbf{inspected\_objects\_json\_format}: \\
\texttt{[\{"object 1" : \{} \\
\texttt{\ \ \ "state":"",} \\
\texttt{\ \ \ "characteristics":"",} \\
\texttt{\ \ \ "additional info":""} \\
\texttt{,\ ...\}]\}} \\

\midrule
\textbf{Prompt}: \\
\{Initial Prompt\} \\
\texttt{<Last 10 logs(from oldest to latest)>} \{last\_10\_history\} \texttt{</Last 10 logs>} \\

\texttt{Here is the definition of information:} \\
\texttt{<Last 10 logs>: Sequence of observation, action-effect, next observation, next action-effect for each turn.} \\

\texttt{Here is the task:} \\
\texttt{Construct your memory about the room, based on the last 10 logs. Follow below guidelines:} \\
\texttt{1. Identify which objects exist on each directional wall, and add the information to "[SPATIAL MEMORY]" section.} \\
\texttt{2. If you have inspected an object via "inspect [object]", you should add the information to "[INSPECTED OBJECTS]" section.} \\
\texttt{3. The objects that you have not inspected via "inspect [object]" should be added to "[UNINSPECTED OBJECTS]" section.} \\
\texttt{4. Add any information from observations which is not included in [SPATIAL MEMORY] and [INSPECTED OBJECTS] that you think is necessary for solving other problems to the "[Additional Memory]" section.} \\

\texttt{Please respond following below format without any other text:} \\
\texttt{[SPATIAL MEMORY]} \{spatial\_json\_format\} \\
\texttt{[INSPECTED OBJECTS]} \{inspected\_objects\_json\_format\} \\
\texttt{[UNINSPECTED OBJECTS] []} \\
\texttt{[ADDITIONAL MEMORY] [1. additional memory1, 2. additional memory2, ...]} \\

\bottomrule
\end{tabular}
\label{tab:experiment-prompts-3}

\end{subfigure}
\label{tab:templatetestend2}

\caption{Prompts for memory management module - for first run (memory construction).}
\label{tab:prompt_trait2}
\end{table*}

\begin{table*}[!p]
\centering
\small

\begin{subfigure}{\textwidth}
\centering
\begin{tabular}{p{0.95\textwidth}}
\toprule
\textbf{Variables}: \\
\texttt{spatial\_memory, action\_history} \\

\midrule
\textbf{Initial prompt}: \\
\texttt{You are an AI agent playing a room escape game. The room is surrounded by 4 walls, and you can explore other walls by "turn\_to\_[direction]". Each wall has objects that you can interact with, and you can inspect the object by "inspect [object]". Please read the given information and task description below carefully and respond accordingly.} \\

\midrule
\textbf{last\_10\_history} (\texttt{for log in action\_history}) : \\
\texttt{Observation: [\{log['scene']\}]\textbackslash n} \\
\texttt{Action: [\{log['action']\}]-\{log['analysis']\}} \\

\midrule
\textbf{spatial\_json\_format}: \\
\texttt{\{"direction 1" : \{} \\
\texttt{\ \ \ "objects":["object1", "object2", ...]} \\
\texttt{\},} \\
\texttt{...\}} \\

\midrule
\textbf{inspected\_objects\_json\_format}: \\
\texttt{[\{"object 1" : \{} \\
\texttt{\ \ \ "state":"",} \\
\texttt{\ \ \ "characteristics":"",} \\
\texttt{\ \ \ "additional info":""} \\
\texttt{,\ ...\}]\}} \\

\midrule
\textbf{Prompt}: \\
\{Initial Prompt\} \\
\texttt{<Current Memory> \{spatial\_memory\} </Current Memory>} \\
\texttt{<Last 10 logs(from oldest to latest)>} \{last\_10\_history\} \texttt{</Last 10 logs>} \\

\texttt{Here is the definition of each information:} \\
\texttt{<Last 10 logs>: Sequence of observation, action-effect, next observation, next action-effect for each turn.} \\

\texttt{Here is the task:} \\
\texttt{Update your memory about the room, based on the last 10 logs. Follow below guidelines:} \\
\texttt{1. If you newly inspected an object via "inspect [object]" among objects in "[UNINSPECTED OBJECTS]" section,} \\
\texttt{\ \ you should add the information to "[INSPECTED OBJECTS]" section.} \\
\texttt{2. Based on the information that you can obtain from last 10 logs, update <Current Memory>} \\
\texttt{\ \ if there exists any information that you can obtain from <Last 10 logs> but not in <Current Memory>.} \\
\texttt{3. Add any information not included in spatial memory and inspected objects that you think is necessary} \\
\texttt{\ \ for solving other problems to the "[ADDITIONAL MEMORY]" section.} \\

\texttt{Please respond following below format without any other text:} \\
\texttt{[SPATIAL MEMORY]} \{spatial\_json\_format\} \\
\texttt{[INSPECTED OBJECTS]} \{inspected\_objects\_json\_format\} \\
\texttt{[UNINSPECTED OBJECTS] []} \\
\texttt{[ADDITIONAL MEMORY] [1. additional memory1, 2. additional memory2, ...]} \\

\bottomrule
\end{tabular}
\label{tab:experiment-prompts-4}

\end{subfigure}
\label{tab:templatetestend3}

\caption{Prompts for Memory management module after memory construction}
\label{tab:prompt_trait3}
\end{table*}

\begin{table*}[!p]
\centering
\small

\begin{subfigure}{\textwidth}
\centering
\begin{tabular}{p{0.95\textwidth}}
\toprule
\textbf{Variables}: \\
\texttt{previous\_scene, previous\_action, current\_scene} \\

\midrule
\textbf{Prompt}: \\
\{Initial Prompt\} \\
\texttt{<Previous Observation> : \{previous\_scene\}} \\
\texttt{<Previous Action> : \{previous\_action\}} \\
\texttt{<Current Observation> : \{current\_scene\}} \\

\texttt{Here is the definition of each information:} \\
\texttt{<Previous Observation> is a description of a scene before your action, and <Current Observation> is a description of a scene after your action.} \\

\texttt{Here is the task:} \\
\texttt{Analyze the effect of your action by comparing <Previous Observation> and <Current Observation>. The analysis should be concise and definitive, not descriptive.} \\
\texttt{Keep it under 10 words.} \\

\texttt{[ANALYSIS]} \\

\bottomrule
\end{tabular}
\label{tab:experiment-prompts-5}

\end{subfigure}
\label{tab:templatetestend4}

\caption{Prompts for Reasoning module-after action.}
\label{tab:prompt_trait4}
\end{table*}

\begin{table*}[!p]
\centering
\small

\begin{subfigure}{\textwidth}
\centering
\begin{tabular}{p{0.95\textwidth}}
\toprule
\textbf{Variables}: \\
\texttt{item\_name} \\

\midrule
\textbf{Prompt}: \\
\texttt{This image is a close-up view of an item '\{item\_name\}'.} \\
\texttt{Describe this image. Your description should fulfill the following rules:} \\
\texttt{1. Description should include every visual information, but concise and clear.} \\
\texttt{2. Do not start description with phrases like 'The image depicts', 'The image shows', etc.} \\

\bottomrule
\end{tabular}
\caption{Prompts for getting caption from image of item-view}
\label{tab:captioning-prompts-1}

\begin{tabular}{p{0.95\textwidth}}
\toprule
\textbf{Variables}: \\
\texttt{object\_type, items\_str} \\

\midrule
\texttt{This image is a close-up view of an object '\{object\_type\}'.} \\
\texttt{In \{object\_type\}, the following objects are present: \{items\_str\}} \\
\texttt{Describe this image. The names of visible objects should be expressed using the given object names above, enclosed in "".} \\
\texttt{Your description should fulfill the following rules:} \\
\texttt{1. Description should include every visual information, but concise and clear.} \\
\texttt{2. Do not include any analysis of the scene or the room, just describe the image.} \\
\texttt{3. Do not start description with phrases like 'The image depicts', 'The image shows', etc.} \\

\bottomrule
\end{tabular}
\caption{Prompts for getting caption from image of object-view}
\label{tab:captioning-prompts-2}

\begin{tabular}{p{0.95\textwidth}}
\toprule
\textbf{Variables}: \\
\texttt{objects} \\

\midrule
\texttt{This image is a wall view of a room, with following objects: \{objects\}.} \\
\texttt{Describe this image. The names of visible objects should be expressed using the given object names above, enclosed in "".} \\
\texttt{Your description should fulfill the following rules:} \\
\texttt{1. Description should include every visual information, but concise and clear.} \\
\texttt{2. Do not include any analysis of the scene or the room, just describe the image.} \\
\texttt{3. Do not start description with phrases like 'The image depicts', 'The image shows', etc.} \\

\bottomrule
\end{tabular}
\caption{Prompts for getting caption from image of wall-view}
\label{tab:captioning-prompts-3}

\end{subfigure}
\label{tab:templatetestend5}

\caption{Prompts for captioning observations.}
\label{tab:prompt_trait5}
\end{table*}

\begin{table*}[!p]
\centering
\small

\begin{subfigure}{\textwidth}
\centering
\begin{tabular}{p{0.95\textwidth}}
\toprule
\textbf{Prompt}: \\
\midrule
\ttfamily
You are a game designer proposing ideas for an 'Escape Room Game'. I want to create an 'Escape Room Game' in a room structure enclosed by four walls. Inside the room, there are various objects, and it's an 'Escape Room Game' where actions initiated from a specific object trigger other objects, allowing puzzles to be solved one by one. There are two types of objects:\vspace{0.5em}

1. Receptacle:\newline
- Definition) Objects that can contain or hold other objects, such as tables, drawers, or wardrobes.\newline
- Each Receptacle is visually distinct and has unique States. For example, a door without a lock has two States: 'open' and 'closed', but a door with a lock has three States: 'locked', 'open', and 'closed'. Also, a drawer with two compartments can have four States depending on the open/closed status of each compartment: 'opened\_opened', 'opened\_closed', 'closed\_opened', 'closed\_closed'.\newline
- Each Receptacle must be assigned to one of the four walls of the room.\newline
- Receptacles can contain Items. However, considering realistic constraints, each Item must define the "Interactable State(s)" of the Receptacle where it can be interacted with (i.e., visible). For example, if you have a key and a wardrobe with States defined as [locked, closed, open], the key's interactable state for that wardrobe is [open]. Interactable States can be multiple. For instance, for a drawer with two compartments, if a key exists in the second compartment, the key's interactable states are [closed\_open, open\_open].\vspace{0.5em}

2. Item:\newline
- Definition) Objects that can be triggers for interaction with specific Receptacles and other Items, such as keys, buttons, photos, etc. Items must be contained within a Receptacle. There are two types of Items:\newline
  - 2-a. Applicable Item: Items that can be picked up, stored in an inventory, and applied to other objects, such as keys or ID cards.\newline
  - 2-b. Quiz Item: Items that contain a puzzle, which, when correctly solved or triggered by a specific action, activates a change in another specific object within the game.\vspace{0.5em}

Furthermore, in addition to objects, the game requires internal logic for escaping the room. For example, a wardrobe is initially locked, but finding and applying a key from elsewhere unlocks it. A drawer is initially locked, but solving a quiz elsewhere opens it. The game logic can be very creative and diverse, but it must be defined according to the following constraints:\vspace{0.5em}

1. It must be limited to logic that changes the state of a Receptacle.\newline
2. Actions related to game logic must be associated with an Item. For example, changing a drawer from a 'locked' State to a 'closed' State via an 'apply Key' action is game logic. However, changing a drawer from an 'opened' State to a 'closed' State via a 'close' action is not game logic.\vspace{0.5em}

If these two constraints are met, the in-game logic can be expressed atomically in the following format:\newline
(receptacle\_name.state\_name\_before\_action, action\_name(item\_name), \\ receptacle\_name.state\_name\_after\_action, checkTrigger(interaction\_name))\vspace{0.5em}

In summary, to design one escape room game, the following game components are needed:\newline
- Approximately 5 Receptacles, all possible states defined for each object, and which wall they belong to.\newline
- Approximately 5 Items, and which Receptacle they belong to.\newline
- Approximately 5 game logic interactions, expressed in the atomic format described above.\vspace{0.5em}

Please design a game logic with following receptacles and items: \newline
[Receptacle] - 2TierDrawer, Cupboard, Safe, Wardrobe, Desk, Door \newline
[Receptacle] - Key1, key2, ID card, button, puzzle

Design an escape room game by creating interactions between objects and puzzles. Storytelling is not required, and the game does not need to be particularly creative or innovative.
\\
\bottomrule
\end{tabular}

\end{subfigure}
\label{tab:templatetestend6}

\caption{Prompts used for game logic design.}
\label{tab:prompt_logicdesign}
\end{table*}

\begin{table*}[!p]
\centering
\small

\begin{subfigure}{\textwidth}
\centering
\begin{tabular}{p{0.95\textwidth}}
\toprule
\textbf{Variables}: \\
\texttt{image1}, \texttt{image2}, \texttt{[Optional]tried\_answers} \\
\midrule
\textbf{\{History\} (if \textit{tried\_answers} is given)}: \\
\texttt{You have tried the following answers: \{tried\_answers\}, but they are all wrong.} \\
\midrule
\textbf{Prompt}: \\
\texttt{\{image1\} is a hint image for the password of the lock described in \{image2\}.} \\
\texttt{You should guess the password of the lock solely based on the hint image.} \\
\texttt{\{History\}} \\
\texttt{Respond ONLY with the your answer, no other text.} \\

\bottomrule
\end{tabular}
\caption{Prompts for multi-image VQA without reasoning.}
\label{tab:multiimage-vqa-prompts-noreasoning}

\begin{tabular}{p{0.95\textwidth}}
\toprule
\textbf{Variables}: \\
\texttt{image1}, \texttt{image2}, \texttt{[Optional]tried\_answers} \\
\midrule
\textbf{\{History\} (if \textit{tried\_answers} is given)}: \\
\texttt{You have tried the following answers: \{tried\_answers\}, but they are all wrong.} \\
\midrule
\textbf{Prompt}: \\
\texttt{\{image1\} is a hint image for the password of the lock described in \{image2\}.} \\
\texttt{You should guess the password of the lock solely based on the hint image.} \\
\texttt{\{History\}} \\
\texttt{Before you guess, think first, and then guess the password. Your thought should be in section [Think], and your answer should be in section [Answer]. In [Answer], respond ONLY with the your answer, no other text.} \\
\texttt{[Think]} \\
\texttt{[Answer]} \\

\bottomrule
\end{tabular}
\caption{Prompts for multi-image VQA with reasoning.}
\label{tab:multiimage-vqa-prompts-reasoning}

\end{subfigure}
\label{tab:templatetestend7}

\caption{Prompts for multi-image VQA.}
\label{tab:prompt_multiimage_vqa}
\end{table*}

\begin{table*}[!p]
\centering
\small

\begin{subfigure}{\textwidth}
\centering
\begin{tabular}{p{0.95\textwidth}}
\toprule
\textbf{Variables}: \\
\texttt{think}, \texttt{action} \\
\midrule
\textbf{Prompt}: \\
\texttt{You are tasked with analyzing the logs of an AI agent in an escape room game.} \\
\texttt{I'll give you a think-action pair that the AI agent has performed.} \\
\texttt{Your task is as follows:} \\
\\
\texttt{1. Divide the agent's reasoning chain into distinct units that fall into the categories explicitly given below:} \\
\texttt{   - Observation: visually perceiving, describing, or analyzing the current scene} \\
\texttt{   - Guess: making a guess or assumption based on priors or knowledge, without exact evidence acquired in the environment} \\
\texttt{   - Hypothesis Formation: forming a hypothesis or idea to test in the environment, based on any evidence acquired in the environment} \\
\texttt{   - Planning: Establishing or formulating action plans within the environment} \\
\texttt{   - Recall: Retrieving information from history - remembering the state/location of objects, actions previously performed, and other relevant details} \\
\\
\texttt{A unit could be:} \\
\texttt{   - A single sentence} \\
\texttt{   - Multiple sentences together} \\
\texttt{   - Multiple units within a single sentence} \\
\\
\texttt{2. If a unit does not fit into the above categories, create a new category called [None-<new category name>] and place it there.} \\
\texttt{3. List all units in the same order as they appear in the original text. Do not reorder or rearrange the units.} \\
\\
\texttt{Input:} \\
\texttt{[Think]\{think\}} \\
\texttt{[Action]\{action\}} \\
\\
\texttt{Response Format:} \\
\texttt{[Category of Unit 1]: Sentence or phrase of given [Think]} \\
\texttt{[Category of Unit 2]: Sentence or phrase of given [Think]} \\
\texttt{[Category of Unit 3]: Sentence or phrase of given [Think]} \\
\texttt{...}
\\
\texttt{Response:} \\

\bottomrule
\end{tabular}

\end{subfigure}
\label{tab:templatetestend8}

\caption{Prompts for annotating components of reasoning.}
\label{tab:prompt_annotating_components}

\end{table*}


\end{document}